%% file: main.tex
\documentclass[10pt,twocolumn,letterpaper]{article}

\usepackage[pagenumbers]{cvpr} % To force page numbers, e.g. for an arXiv version

\usepackage[dvipsnames]{xcolor}

\usepackage{multirow}
\usepackage{colortbl}
\usepackage{soul}
\usepackage[pagebackref,breaklinks,colorlinks,citecolor=cvprblue]{hyperref}
\usepackage{pifont}
\usepackage{longtable}
\usepackage{array}
\usepackage{framed}
\usepackage{wrapfig}

\def\thickhline{\noalign{\hrule height1.2pt}}

\newcommand{\om}{GSVA}
\newcommand{\omlong}{Generalized Segmentation Vision Assistant}
\newcommand{\tabincell}[2]{\begin{tabular}{@{}#1@{}}#2\end{tabular}}

\definecolor{mygray}{gray}{.9}
\definecolor{cvprblue}{rgb}{0.21,0.49,0.74}

\def\paperID{10060} % *** Enter the Paper ID here
\def\confName{CVPR}
\def\confYear{2024}

\title{GSVA: Generalized Segmentation via Multimodal Large Language Models}

\author{
  Zhuofan~Xia\thanks{Equal contribution.}\hspace{12pt}
  Dongchen~Han\footnotemark[1]\hspace{12pt}
  Yizeng~Han\hspace{12pt}
  Xuran~Pan\hspace{12pt}
  Shiji~Song\hspace{12pt}
  Gao~Huang\thanks{Corresponding author.}\\
  \normalsize{Department of Automation, BNRist, Tsinghua University}
}

\begin{document}
\maketitle
\input{sec/abstract}    
\input{sec/introduction}
\input{sec/relatedworks}
\input{sec/method}
\input{sec/experiments}
\input{sec/conclusion}
% \newpage
\input{sec/acknowledgements}

\clearpage
\appendix
\section*{Appendix}
\input{sec/discussion}

\input{sec/impl}
\input{sec/refzom}
\input{sec/semseg}
\input{sec/abl_ana}
\input{sec/visualizemore}
\clearpage
{
    \small
    \bibliographystyle{ieeenat_fullname}
    \bibliography{main}
}

\end{document}

%% file: sec/abstract.tex
\begin{abstract}

Generalized Referring Expression Segmentation (GRES) extends the scope of classic RES to refer to multiple objects in one expression or identify the empty targets absent in the image. GRES poses challenges in modeling the complex spatial relationships of the instances in the image and identifying non-existing referents. Multimodal Large Language Models (MLLMs) have recently shown tremendous progress in these complicated vision-language tasks. Connecting Large Language Models (LLMs) and vision models, MLLMs are proficient in understanding contexts with visual inputs. Among them, LISA, as a representative, adopts a special [SEG] token to prompt a segmentation mask decoder, e.g., SAM, to enable MLLMs in the RES task. However, existing solutions to GRES remain unsatisfactory since current segmentation MLLMs cannot correctly handle the cases where users might reference multiple subjects in a singular prompt or provide descriptions incongruent with any image target. In this paper, we propose Generalized Segmentation Vision Assistant (GSVA) to address this gap. Specifically, GSVA reuses the [SEG] token to prompt the segmentation model towards supporting multiple mask references simultaneously and innovatively learns to generate a [REJ] token to reject the null targets explicitly. Experiments validate GSVA's efficacy in resolving the GRES issue, marking a notable enhancement and setting a new record on the GRES benchmark gRefCOCO dataset. GSVA also proves effective across various classic referring segmentation and comprehension tasks. 
Code is available at \url{https://github.com/LeapLabTHU/GSVA}.

\end{abstract}

%% file: sec/introduction.tex
\section{Introduction}
\label{sec:intro}

\begin{figure}
\centering
\includegraphics[width=\linewidth]{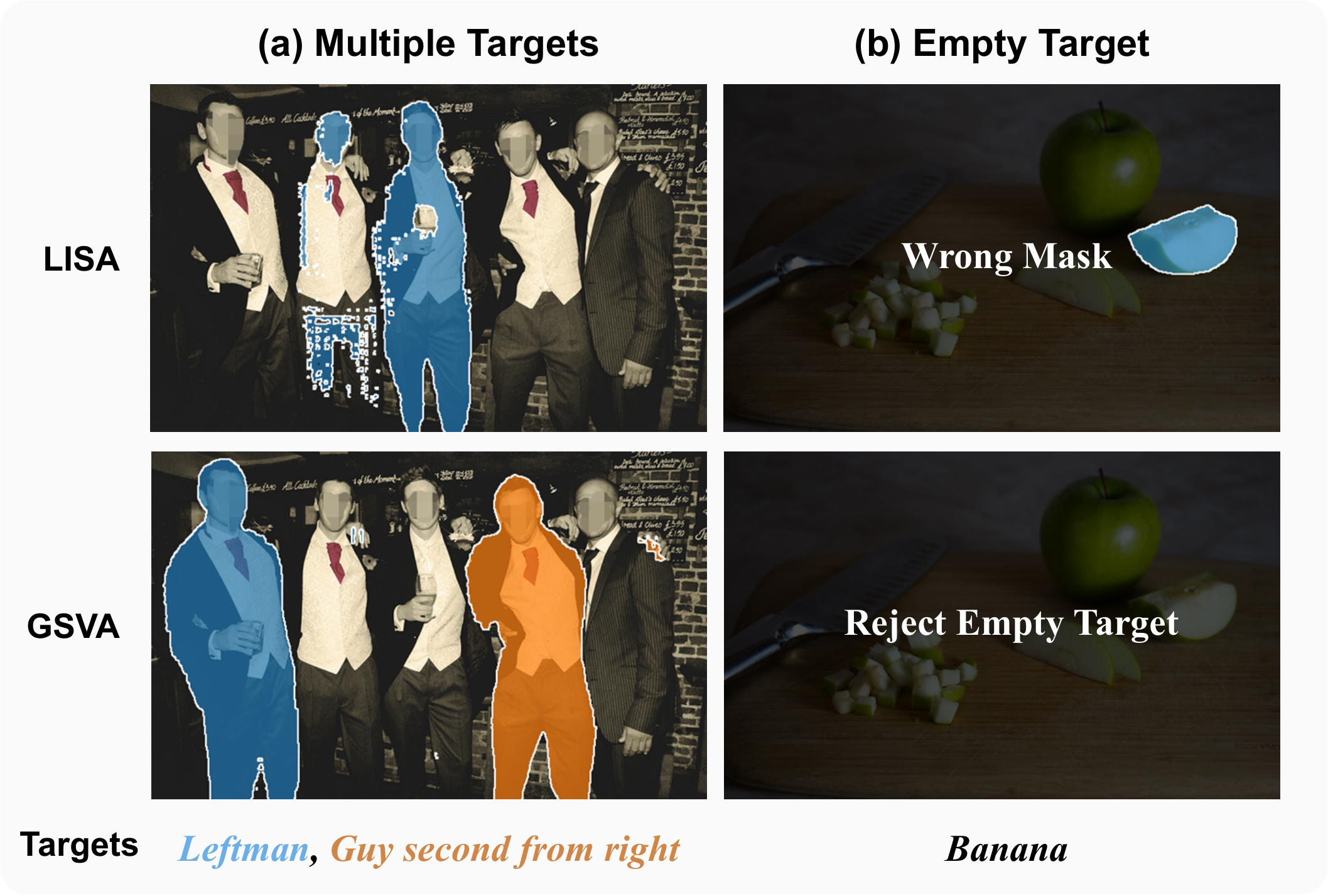}
\vspace{-0.3in}
\caption{Comparison of the segmentation masks by LISA~\cite{lai2023lisa} and \om{}, facing the challenges in Generalized Referring Expression Segmentation (GRES)~\cite{liu2023gres}. (a) LISA fails to segment the correct targets when multiple targets are requested due to the
single [SEG] token restriction. \om{} successfully generates all target masks via learning multiple [SEG] tokens. (b) When the referent does not exist in the image, \textit{i.e.}, the empty target is requested, LISA reluctantly produces the wrong mask because of the compulsive [SEG] token output. In contrast, \om{} can reject the empty targets by predicting [REJ] tokens in the output sequence.}
\label{fig:fig1}
\vspace{-0.25in}
\end{figure}

% 从RES开始，引出RES的局限性
Referring Expression Segmentation (RES)~\cite{cheng2014imagespirit,hu2016segmentation} is an emerging vision-language (VL) task predicting the masks of the interested objects referred to in the language expression. RES has great potential in many areas, especially embodied AI~\cite{wang2019reinforced,qi2020reverie,gao2021room,sima2023embodied,huang2024joint}, including VL navigation, human-robot interaction, \textit{etc.} Nevertheless, the simplification in RES formulation that one referring expression must match an individual object in the image~\cite{kazemzadeh2014referitgame} has left a gap between current RES algorithms and real-world applications, neglecting multiple-target and empty-target cases.

% 点出两类问题，给出GRES
To bridge this gap, Generalized Referring Expression Segmentation (GRES)~\cite{liu2023gres} has recently been proposed to support multiple-target and empty-target cases. Practically, users refer to multiple subjects within a single instruction or provide descriptions that do not correspond to any targets in the image. As an example shown in Figure~\ref{fig:fig1} (a), the left man and the second man from the right are targeted simultaneously, while a banana is referred to in a scene of apples in Figure~\ref{fig:fig1} (b). RES takes no account of these cases, which GRES handles. In addition to the multimodal correspondences between images and text prompts in classic RES, GRES poses new challenges in handling more complicated multiple-target and empty-target cases. Therefore, the models must handle complex spatial relationships of the instances in the image~\cite{liu2023gres} to segment the targets at various locations and reject the empty targets in the wrong places.

% 指出GRES的挑战性，建模复杂的空间关系，引出MLLM
The recent blooming Multimodal Large Language Models (MLLMs)~\cite{alayrac2022flamingo,zhang2023lladp,ye2023mplug,liu2023llava,li2023blip2} meet the requirements of GRES since they show excellency in complex reasoning~\cite{chung2022scaling} and instruction following~\cite{ouyang2022training} with visual inputs by aligning the LLMs~\cite{brown2020language,openai2023gpt4,touvron2023llama,touvron2023llama2,alpaca2023,vicuna2023} and Visual Foundation Models (VFMs)~\cite{radford2021learning,wang2023image,fang2023eva,xiao2023florence} which are typically various Vision Transformers~\cite{dosovitskiy2020image,xia2022vision,xia2023dat++,pan2022integration,pan2023slide,han2023agent,han2023dynamic} to perceive image or video inputs. To support segmentation output, many works~\cite{zhang2023next,wu2023vstar,wu2023see} link an MLLM (\textit{e.g.}, LLaVA~\cite{liu2023llava}) and a segmentation model (\textit{e.g.}, SAM~\cite{kirillov2023segment}) by prompting the decoder with special token embeddings (\textit{e.g.}, [SEG] in LISA~\cite{lai2023lisa}) to generate masks of the referents in the user's instructions. Although these models manage to handle RES, GRES is still beyond their reach. As shown in Figure~\ref{fig:fig1}, LISA fails to work well in GRES where the multiple-target and empty-target challenges remain uncharted.

To address the above challenges, we propose \textbf{\omlong{} (\om{})}. We attribute the vulnerability of other segmentation MLLMs in GRES to the single constant [SEG] token that restricts its flexibility. Therefore, we present two pivotal designs in \om{}: (1) learning to predict multiple [SEG] tokens to segment multiple targets; (2) rejecting empty targets in referring expressions by predicting [REJ] tokens. Specifically, when multiple targets are requested in the referring expression, we place multiple weight-sharing [SEG] tokens corresponding to the entities in the expression, encouraging the MLLM to learn to output multiple [SEG] tokens. To distinguish each [SEG] token and avoid ambiguity, we add the expression of each entity in front of the corresponding [SEG] token, hinting each [SEG] token to focus on the specific target, which can be regarded as implicit In-Context Learning, and dynamic neural network~\cite{han2021dynamic}. Meanwhile, if the referents are absent in the image, the corresponding [SEG] tokens after the prompts are altered to [REJ] tokens to identify empty targets. The predicted [REJ] tokens are directly assigned with empty masks without decoded, which liberates the segmentation model from seeking non-existing targets in the image. This Benefiting from these novel designs, \om{} takes a big step forward in addressing GRES challenges, as shown in the second row of Figure~\ref{fig:fig1}.

Our contributions are summarized as follows: (1) We propose \om{} to solve the GRES problem with MLLM by handling the spatial relationships among targets, and study the GRES problem systematically in the context of LLM for the first time. (2) We propose the non-trivial \textbf{shared-weight multiple [SEG] tokens guided by each referent prompt} to address the multiple-target problem. (3) We firstly propose a clean solution, \textbf{the [REJ] token}, to reject the empty targets, which can be seamlessly applied to various models. (4) \om{} is intuitive and effective, \textbf{achieving state-of-the-art performance on the GRES benchmarks}.

%% file: sec/relatedworks.tex
\section{Related Works}
\label{sec:related}

\noindent
\textbf{RES and GRES.} Referring Expression Segmentation (RES) \cite{cheng2014imagespirit,hu2016segmentation,chen2021yourefit,mao2016generation} assumes that one expression matches one existing target, and many works explore fusing image and language~\cite{liu2017recurrent,li2018referring,chen2019see,feng2021encoder,yang2022lavt} to segment objects under instructions. Currently, most RES methods adopt the cross-attention module or cross-modal alignment to bridge the modality gap~\cite{shi2018key,ye2019cross,zhang2022coupalign,wei2023learning,chng2023mask}. Another line of research enables the text prompts for segmentation model with a unified decoder~\cite{kirillov2023segment, zou2023generalized, zou2023segment, liu2023grounding}, offering more flexible outputs. To break the jail for the arbitrary number of targets, DMMI~\cite{hu2023beyond} focuses on the one-to-many setting where text expression refers to varying numbers of targets.ReLA~\cite{liu2023gres} proposes the Generalized Referring Expression Segmentation (GRES) task, supporting both the multi-target and empty-target scenarios, which is our main research scope.

\noindent
\textbf{MLLM.} Multimodal Large Language Models (MLLMs) align the vision and language modalities by various techniques, including cross-attention module~\cite{alayrac2022flamingo}, prompt tuning tokens~\cite{zhang2023lladp}, Q-Former~\cite{li2023blip2,dai2023instruct}, linear projection layers~\cite{liu2023llava}, and unified model architectures~\cite{peng2023kosmos,xiao2023florence}. Endowed with unprecedented capabilities in reasoning with world-knowledge and following instructions of users, MLLM shows extraordinary performances in various vision language tasks~\cite{lv2023kosmos,wang2023cogvlm}. Equipped with the segmentation decoders or detection heads, MLLMs can also excel in the vision-centric tasks, such as object detection and segmentation~\cite{wang2023visionllm, zhang2023next, xu2023u, rasheed2023glamm, lin2023sphinx}. Among them, LISA~\cite{lai2023lisa} makes the most of the reasoning ability with a [SEG] token to address the Reasoning Segmentation problem. However, LISA fails to tackle the challenge in GRES due to the inflexible [SEG] token, which is addressed by our proposed \om{}.

\noindent
\textbf{Dynamic Networks.} Dynamic neural networks~\cite{han2021dynamic} can adapt their architectures~\cite{huang2022glance,wang2021not,wang2021adaptive,wang2022adafocus,wang2022adafocusv3,han2022learning,han2023latency,han2021spatially,han2022latency} or parameters~\cite{pu2023adaptive,ding2021repvgg,ding2022scaling} to different inputs or switch the computation architecture in adjustment to different different time steps~\cite{xia2023budgeted,wang2023efficienttrain,ni2023deep,ni2024revisit}, in order to achieve better accuracy and efficiency. In \om{}, the weight-sharing [SEG] tokens adapt to multiple targets under the hint of the prepended target prompts and dynamically reject empty targets with an individual prediction of rejection tokens. 

%% file: sec/method.tex
\section{\omlong{}}
\label{sec:method}

\begin{figure*}
    \centering
    \includegraphics[width=\linewidth]{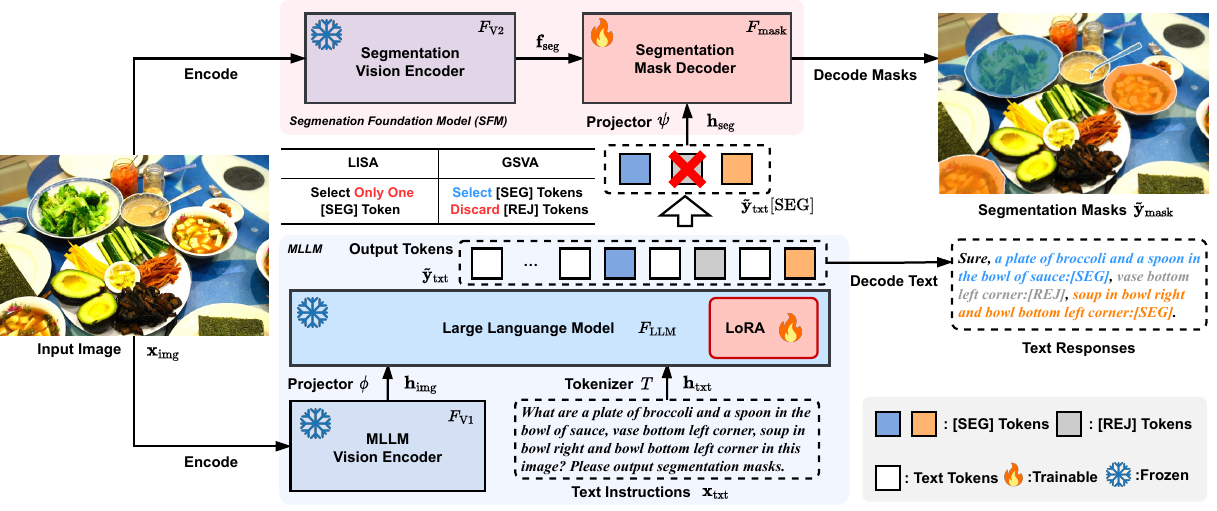}
    \vspace{-0.2in}
    \caption{Overview of \om{}. At the bottom of the figure, the MLLM encodes the input image and concatenates the tokenized text tokens to follow instructions. \om{} generates multiple [SEG] tokens to handle multiple referred targets and rejects the objects absent in the image through [REJ] tokens. At the top of the figure, the SFM also encodes the image for segmentation and selects all [SEG] tokens in the output sequence to prompt the mask decoder to segment the target objects referred to in the instructions.}
    \label{fig:fig2}
    \vspace{-0.2in}
\end{figure*}

In this section, we initiate with the introduction of the model design of \textbf{\omlong{}} (\textbf{\om{}}), which is followed by the analysis of some certain limitations of LISA in Generalized Referring Segmentation (GRES). Subsequently, we delve into the introduction of two pivotal elements of \om{}, segmenting multiple targets and learning the rejection token. These components are fundamental in the conceptualization and construction of \om{}.

\subsection{Model Architecture}

The architecture of \om{} is illustrated in Figure~\ref{fig:fig2}, resembling LISA~\cite{lai2023lisa}, which enables high-fidelity segmentation outputs by integrating two types of foundation models: (1) Multimodal Large Language Model (MLLM) as an aligned vision-language cognitive module; (2) Segmentation Foundation Model (SFM) to segment the target out of the input image based on user's instruction. To connect these two modules, LISA proposes a paradigm named \textit{embedding as mask} where an extra [SEG] token is appended to the vocabularies of the MLLM and serves as the prompt of the SFM to segment the target following the intention of the user.

\noindent
\textbf{Multimodal Large Language Model.} The MLLM consists of a decoder-based language model $F_\text{LLM}$ to auto-regressively generate text responses following the user's inputs, a vision encoder $F_\text{V1}$ to extract features from the input image, and a linear projector $\phi$ to align the representations between image and text modalities. Specifically, the pretrained LLaVA~\cite{liu2023llava} variants with CLIP-ViT-L/14~\cite{radford2021learning} and Vicuna-7B/13B~\cite{vicuna2023} are employed. Given an input image $\mathbf{x}_\text{img}$, the vision encoder $F_\text{V1}$ first encodes it into image features, and then the projector $\phi$ maps the features into the visual token embeddings in the LLM input space:
\vskip -0.15in
\begin{equation}
    \mathbf{h}_\text{img}=\phi(F_\text{V1}(\mathbf{x}_\text{img})),
    \label{eq:get_himg}
\end{equation}
\vskip -0.05in
where the input image $\mathbf{x}_\text{img}$ is typically resized to $h\!\times{}\!w\!\times{}\!3$, and the image tokens $\mathbf{h}_\text{img}\in\mathbb{R}^{n_\text{img}\!\times{}\!d}$ is aligned with the language modality. For CLIP-ViT-L/14, the input image with $h\!=\!w\!=\!224$ is encoded with ViT of patch size in 14, therefore the length of tokens $n_\text{img}\!=\!hw/14^2\!=\!256$, and the LLM dimensions $d$ are 4096 and 5120 for Vicuna-7B/13B, respectively. Along with the input image, the text instructions describing the targets to segment are tokenized into text tokens by the LLM tokenizer $T$:
\vskip -0.15in
\begin{equation}
    \mathbf{h}_\text{txt}=T(\mathbf{x}_\text{txt}).
    \label{eq:get_htxt}
\end{equation}
\vskip -0.05in
The image tokens and text tokens are concatenated together and then fed into the LLM after prepending a sequence of fixed prompt tokens $\mathbf{h}_\text{prompt}$ (omitted in the figure). The output token embeddings $\tilde{\mathbf{y}}_\text{txt}$ are generated auto-regressively:
\vskip -0.1in 
\begin{equation}
\tilde{\mathbf{y}}_\text{txt}=F_\text{LLM}\left([\mathbf{h}_\text{prompt}||\mathbf{h}_\text{img}||\mathbf{h}_\text{txt}]\right),
    \label{eq:llm_output}
\end{equation}
\vskip -0.05in
where $||$ is concatenation operation. The text responses are obtained from $\tilde{\mathbf{y}}_\text{txt}$ by applying a linear classifier to predict the next words in the vocabulary. 

In LISA, a special token [SEG] is appended in the vocabulary to activate the segmentation ability of MLLM. The model learns to predict [SEG] token in the output sequence to indicate there is a target to segment. LISA then selects the output embedding of the [SEG] token $\tilde{\mathbf{y}}_\text{txt}[\text{SEG}]$, and projects it into the prompt space of the SFM by an MLP projector $\psi$: 
\vskip -0.2in
\begin{equation}
    \mathbf{h}_\text{seg}=\psi\left(\tilde{\mathbf{y}}_\text{txt}[\text{SEG}]\right).
    \label{eq:get_hseg}
\end{equation}
\vskip -0.05in
The segmentation model is hence ready to decode the target mask from the query token $\mathbf{h}_\text{seg}$.

\noindent
\textbf{Segmentation Foundation Model.} The SFM is a query-based segmentation model, where a frozen vision encoder $F_\text{V2}$ takes in images of higher resolution than the vision encoder in MLLM to keep more details, followed by a trainable mask decoder $F_\text{mask}$ to decode masks from the queries. The pretrained SAM~\cite{kirillov2023segment} with ViT-H backbone is instantiated as the SFM to produce high-quality masks. The given input image $\mathbf{x}_\text{img}$ is resized to a larger resolution $H\!\times{}\!W\!\times{}\!3$, with $H\!=\!W\!=\!1024$ and encoded with $F_\text{V2}$ to extract features $\mathbf{f}_\text{seg}$ for segmentation:
\vskip -0.15in
\begin{equation}
    \mathbf{f}_\text{seg}=F_\text{V2}(\mathbf{x}_\text{img}).
    \label{eq:get_fseg}
\end{equation}
\vskip -0.05in
Condition on the features $\mathbf{f}_\text{seg}\in\mathbb{R}^{\frac{H}{16}\!\times{}\!\frac{W}{16}\!\times{}\!C}$ with $C\!=\!256$ for SAM, the mask decoder $F_\text{mask}$ decodes the segmentation masks from the segmentation queries $\mathbf{h}_\text{seg}\in\mathbb{R}^{N_\text{seg}\times{}C}$:
\vskip -0.15in
\begin{equation}
    \tilde{\mathbf{y}}_\text{mask}=F_\text{mask}(\mathbf{h}_\text{seg}|\mathbf{f}_\text{seg}),
    \label{eq:mask_decoder}
\end{equation}
\vskip -0.05in
where each query in $\mathbf{h}_\text{seg}$ corresponds to one segmentation mask in $\tilde{\mathbf{y}}_\text{mask}$.

LISA assumes that only one target exists to segment in the input image and its corresponding instructions. However, in \om{}, we extend it to a new scenario with multiple targets and empty targets, including multiple [SEG] tokens to invoke segmentation and [REJ] tokens to reject unreasonable instructed targets absent in the image. As shown in Figure~\ref{fig:fig2}, \om{} supports multiple [SEG]/[REJ] tokens in the output sequence and selects all the [SEG] tokens and discards every [REJ] token after attaining $\tilde{\mathbf{y}}_\text{txt}$ in Eq.~\eqref{eq:llm_output}. Therefore, there is more than one query in $\mathbf{h}_\text{seg}$, thus enabling the SFM to segment multiple targets. These designs make \om{} competent in the GRES task, where the awareness of multiple and empty targets is of vital importance.

\subsection{GRES: Task and Challenges}

\noindent
\textbf{Task.} Generalized Referring Expression Segmentation (GRES)~\cite{liu2023gres} removes the constraint on the number of referred targets in the expression in the conventional Referring Expression Segmentation (RES)~\cite{kazemzadeh2014referitgame,mao2016generation,yu2016modeling}. Different from that one expression only refers to one instance or region in RES, GRES allows arbitrary numbers of referred targets, including multiple instances or no target circumstances. In GRES, the user can refer to many instances simultaneously or include the objects that do not exist in the image. For instance, there are three referring expressions in Figure~\ref{fig:fig2}, including \textit{a plate of broccoli and a spoon in the bowl of sauce}, \textit{vase bottom left corner}, and \textit{soup in bowl right and bowl bottom left corner}. In the GRES case, the model ought to segment the masks of the objects referred to in the 1$^\text{st}$ and 3$^\text{rd}$ expression, meanwhile producing an empty mask for the 2$^\text{nd}$ expressions since there are no vases present at the bottom left corner.

\noindent
\textbf{Challenges.} The challenges in GRES are common in practice, especially in embodied AI~\cite{wang2019reinforced,qi2020reverie,gao2021room,sima2023embodied}. The one challenge is \textbf{multiple targets}. Take robot navigation and planning~\cite{nguyen2019vision,wang2019reinforced,chen2021yourefit} as an example, a robot may be asked to perceive multiple targets in the surrounding environment, e.g., to \textit{bring the two bowls of soup} in Figure~\ref{fig:fig2}. The vision system of the robot needs to locate and segment the containers holding the referred food one at a time. The other challenge is \textbf{empty target}. Suppose the robot is ordered to cut an apple with a knife in the scene of Figure~\ref{fig:fig2}, whereas no apple is in the view of the camera, the vision system has to identify that the referred object is not in the scene. If it relies on some conventional RES methods which assume the expression must match something in the image, the output of the vision system could be undefined and potentially dangerous in some real-world cases.

\noindent
\textbf{Differences from ReasonSeg.} Reasoning Segmentation (ReasonSeg) proposed by LISA~\cite{lai2023lisa} emphasizes the complex text instructions in RES. In ReasonSeg, the instructions are more implicit and sophisticated, forcing models to reason using world knowledge. Besides, the logic chain is usually longer and more challenging in ReasonSeg, which requires the model to deduce the final target object referred to in the image. In contrast, GRES increases complexity in another dimension by involving complicated spatial relationships. Hence, the model has to learn to handle this spatial information and understand the relationships between the instances. To meet these requirements, LISA tunes MLLMs with complex instructions paired with masks, while \om{} arouses the spatial modeling capabilities of MLLMs by learning multiple targets and rejecting empty targets.
 
\subsection{Multiple [SEG] Tokens for Multiple Targets}

\noindent
\textbf{Single [SEG] token.} LISA~\cite{lai2023lisa} follows the classic RES methods to generate a segmentation mask under given instructions by adding the [SEG] token into the answer. The prompt is formatted as:
\vspace{-0.1in}
\begin{leftbar}
\noindent
{\small \textit{\textbf{User:} What is \{obj\} in this image? Please output segmentation mask.}}
\noindent
{\small \textit{\textbf{Assistant:} Sure, it is [SEG].}}
\end{leftbar}
\vspace{-0.1in}

\noindent
In the above prompt, \textit{\{obj\}} represents an instance referred to or some semantic area to segment, and the embeddings of the [SEG] token output from the LLM are projected to prompt the SFM. When multiple instances are requested simultaneously, this prompt would confuse the model since only one [SEG] token is forced to match several targets. As shown in Figure~\ref{fig:fig1} (a), LISA coercively predicts the masks of the left man and the guy second from the right, leading to a tattered mask and masking the wrong location.

\begin{figure}
\centering
\includegraphics[width=\linewidth]{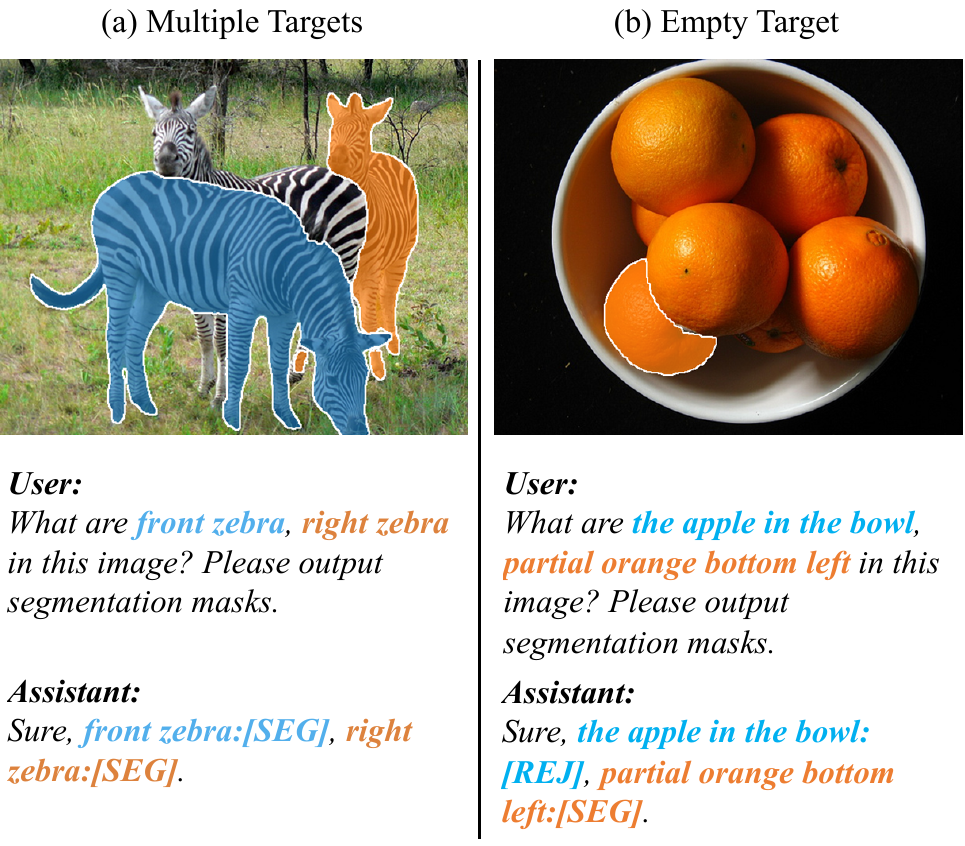}
\vspace{-0.3in}
\caption{Example of the prompts and predicted masks of \om{}-Vicuna-7B drawn from gRefCOCO validation set. (a) depicts the multiple-target case, in which two zebras referred to are handled with two separate [SEG] tokens. (b) shows the empty-target case, where no apple is in the bowl. Thus, the null referent is rejected with a [REJ] token, and no segmentation mask will be generated.}
\label{fig:fig3}
\vspace{-0.25in}
\end{figure}

\noindent
\textbf{Multiple [SEG] tokens.} To mitigate this issue in GRES, we relax this constraint for \om{} to support multiple target outputs via learning multiple [SEG] tokens. To avoid the ambiguity between the [SEG] tokens and the corresponding objects, we prepend the referring expression before each [SEG] token, \textit{i.e.}, each segmentation prompt in the output text is \textit{\{obj\_n\}:[SEG]}, as shown in Figure~\ref{fig:fig3} (a). Streaming all the pairs of object and [SEG] token into one sentence, the question-answer prompt is formatted as (\textit{e.g.}, two targets):

\vspace{-0.1in}
\begin{leftbar}
\noindent
{\small \textit{\textbf{User:} What are \{obj1\}, \{obj2\} in this image? Please output the segmentation masks.}}

\noindent
{\small \textit{\textbf{Assistant:} Sure, \{obj1\}:[SEG], \{obj2\}:[SEG].}}
\end{leftbar}
\vspace{-0.1in}
\noindent
% This prompt requests the MLLM to identify and distinguish different objects in the image based on the instructions and infuse the corresponding location information of each target into the associated [SEG] token.
This prompt requests the MLLM to identify and distinguish different objects in the image based on the instructions and infuse the corresponding location information of each target into the associated [SEG] token. We regard this ability as an implicit multimodal version of In-Context Learning (ICL), which is demonstrated by many prior works~\cite{brown2020language,dong2022survey,zhao2023mmicl,tai2023link,li2023mimicit}. \om{} takes the target description preceding each [SEG] token as a hint to link this token to the object requested in the image through auto-regressive decoding. 

\subsection{Rejecting Empty Targets via [REJ] Tokens}

\noindent
\textbf{Empty targets.} In GRES, many expressions match no targets in the image, including absence, incorrect attributes, inaccurate locations, \textit{etc.} These expressions should be treated as empty targets for models to predict all-negative masks. LISA~\cite{lai2023lisa} falls short of predicting masks with all-zeros seamlessly since the [SEG] token always calls for a segmentation mask. As shown in Figure~\ref{fig:fig1} (b), LISA incorrectly marks a piece of green apple as the empty target banana.

\noindent
\textbf{[REJ] token.}  We let the MLLM of \om{} predict a [REJ] token for each object that does not exist in the image but is referred to in the instructions, as shown in Figure~\ref{fig:fig3} (b). \om{} predicts the targets marked with [REJ] tokens as empty targets, therefore setting all-zero masks for them. The involvement of [REJ] tokens directly rejects the empty target, liberating the mask decoder of the SFM from learning to segment the inexistent targets. An example prompt with one existing target and one empty target is as follows:

\vspace{-0.1in}
\begin{leftbar}
\noindent
{\small \textit{\textbf{User:} What are \{obj1\} (\textbf{absent}), \{obj2\} (\textbf{absent}), \{obj3\} in this image? Please output the segmentation masks.}}

\noindent
{\small \textit{\textbf{Assistant:} Sure, \{obj1\}:[REJ], \{obj2\}:[REJ], \{obj3\}:[SEG].}}
\end{leftbar}
\vspace{-0.1in}

\noindent
The [REJ] token prediction can also be seen as a variant of VQA task, where the specified object and its position in the image need to be considered. Thanks to the unprecedented capabilities of MLLM in understanding the images~\cite{liu2023llava,zhu2023minigpt,li2023blip2,dai2023instruct} and reasoning the spatial relationships of the referring objects~\cite{peng2023kosmos,zhang2023gpt4roi,zhao2023bubogpt}, we make the most of the MLLM in \om{} to unleash the burden of the segmentation model. The proposed empty-target-aware mechanism both improves the quality of masks and ameliorates the errors of identifying nonexistent objects.

%% file: sec/experiments.tex
\section{Experiments}
\label{sec:exp}

In this section, we conduct comprehensive experiments to validate the efficacy of \om{}. First, we show the results on gRefCOCO~\cite{liu2023gres} dataset to show the superiority of \om{} in GRES tasks. Then we verify \om{} that is also competent with other baselines in classic RES, REC tasks. We move on to ablate some important design choices of \om{}, followed by some qualitative visualization of GRES results.

\begin{table*}
\begin{center}
\setlength{\tabcolsep}{4.5mm}{
\renewcommand\arraystretch{1.0}
\resizebox{\linewidth}{!}{
\begin{tabular}{l|ccc|ccc|ccc}
\toprule
\multicolumn{10}{c}{\textbf{Generalized Referring Expression Segmentation on gRefCOCO}} \\
\multirow{2}{*}{Method} & \multicolumn{3}{c|}{Validation Set} & \multicolumn{3}{c|}{Test Set A}  & \multicolumn{3}{c}{Test Set B} \\
& gIoU & cIoU & N-acc. & gIoU & cIoU & N-acc. & gIoU & cIoU & N-acc. \\
\midrule
\midrule
MattNet~\cite{yu2018mattnet} & 48.24 & 47.51 & 41.15 & 59.30 & 58.66 & 44.04 & 46.14 & 45.33 & 41.32 \\
LTS~\cite{jing2021locate} & 52.70 & 52.30 & - & 62.64 & 61.87 & - & 50.42 & 49.96 & - \\
VLT~\cite{ding2021vision} & 52.00 & 52.51 & 47.17 & 63.20 & 62.19 & 48.74 & 50.88 & 50.52 & 47.82 \\
CRIS~\cite{wang2022cris} & 56.27 & 55.34 & - & 63.42 & 63.82 & - & 51.79 & 51.04  & - \\
LAVT~\cite{yang2022lavt} & 58.40 & 57.64 & 49.32 & 65.90 & 65.32 & 49.25 & 55.83 & 55.04 & 48.46 \\
ReLA~\cite{liu2023gres} & 63.60 & 62.42 & 56.37 & 70.03 & 69.26 & 59.02 & 61.02 & 59.88 & 58.40 \\
\midrule
LISA-Vicuna-7B~\cite{lai2023lisa} & 32.21 & 38.72 & 2.71 & 48.54 & 52.55 & 6.37 & 39.65 & 44.79 & 5.00 \\
\rowcolor{mygray}
\om{}-Vicuna-7B & 63.32 & 61.70 & 56.45 & 70.11 & 69.23 & 63.50 & 61.34 & 60.26 & 58.42  \\
LISA-Vicuna-7B~\cite{lai2023lisa} (ft) & 61.63 & 61.76 & 54.67 & 66.27 & 68.50 & 50.01 & 58.84 & \textbf{60.63} & 51.91 \\
\rowcolor{mygray}
\om{}-Vicuna-7B (ft) & \textbf{66.47} & \textbf{63.29} & \textbf{62.43} & \textbf{71.08} & \textbf{69.93} & \textbf{65.31} & \textbf{62.23} & 60.47 & \textbf{60.56} \\
\midrule
LISA-Vicuna-13B~\cite{lai2023lisa} & 32.73 & 39.85 & 3.66 & 48.76 & 53.62 & 4.89 & 39.49 & 45.35 & 4.41 \\
\rowcolor{mygray}
\om{}-Vicuna-13B & 62.97 & 60.18 & 58.44 & 67.17 & 67.59 & 54.60 & 58.06 & 57.28 & 52.22 \\
LISA-Vicuna-13B~\cite{lai2023lisa} (ft) & 63.45 & 62.99 & 55.25 & 68.18 & 69.65 & 52.16 & 61.84 & \textbf{62.24} & 56.15 \\
\rowcolor{mygray}
\om{}-Vicuna-13B (ft) & \textbf{68.01} & \textbf{64.05} & \textbf{65.36} & \textbf{71.75} & \textbf{70.51} & \textbf{67.25} & \textbf{63.83} & 61.28 & \textbf{63.11} \\
\midrule
LISA-Llama2-13B~\cite{lai2023lisa} & 33.26 & 39.64 & 3.27 & 49.76 & 53.80 & 7.28 & 40.49 & 45.41 & 5.73 \\
\rowcolor{mygray}
\om{}-Llama2-13B & 63.20 & 62.38 & 54.51 & 69.52 & 69.86 & 57.84 & 62.06 & 60.77 & 58.30 \\
LISA-Llama2-13B~\cite{lai2023lisa} (ft) & 65.24 & 63.96 & 57.49 & 69.99 & 71.00 & 55.43 & 62.11 & 62.29 & 56.34 \\
\rowcolor{mygray}
\om{}-Llama2-13B (ft) & \textbf{70.04} & \textbf{66.38} & \textbf{66.02} & \textbf{73.29} & \textbf{72.79} & \textbf{64.72} & \textbf{65.45} & \textbf{63.20} & \textbf{62.47} \\
\bottomrule
\end{tabular}}
}
\end{center}
\vspace{-0.2in}
\caption{Generalized referring expression segmentation (GRES) results on gRefCOCO~\cite{liu2023gres} dataset. gIoU and cIoU are IoU metrics averaged by each example and accumulated over whole dataset, respectively. N-acc. is short for the accuracy of correctly classifying null targets. Baselines are copied from~\citet{liu2023gres}. (ft) denotes the model is finetuned on the training set of gRefCOCO.}
\label{tab:gres}
\vspace{-0.25in}
\end{table*}

\subsection{GRES}

\noindent
\textbf{Settings.} We adopt gRefCOCO~\cite{liu2023gres} dataset to validate \om{} and LISA~\cite{lai2023lisa} on GRES, which contains 278,232 expressions, including 80,022
multi-target and 32,202 empty-target ones, referring to the objects in 19,994 images. The images are split into four subsets: training, validation, test-A, and test-B, following the same UNC partition of RefCOCO~\cite{yu2016modeling}. We first add the gRefCOCO training set into the mixed training dataset in LISA to pretrain \om{} and LISA for 50,000 steps and then finetune the models on the gRefCOCO training dataset for another 10 epochs. We evaluate the pretrained and finetuned models on the remaining validation set, test set A, and test set B, respectively. We adopt the gIoU, cIoU metrics for the segmentation mask outputs. Following the implementation in~\citet{liu2023gres}, gIoU averages the IoU for each mask, whereas cIoU computes the cumulative intersection area over the cumulative union area across the whole dataset. As for the empty target, we compute the No-target-accuracy (N-acc.), which is the ratio of the correctly classified empty-target expressions over all the empty-target expressions in the dataset. For a correctly classified empty target, the gIoU is set to 1.0, and the cIoU does not take them into account, while for a misclassified empty target, the gIoU is set to 0.0 and the union area is accumulated in the cIoU. Following the criteria in~\citet{liu2023gres}, a predicted mask is regarded as empty for LISA if the positive pixels are less than 50, whereas \om{} predicts [REJ] tokens to identify empty targets.

\noindent
\textbf{Results.} We report the GRES segmentation results of \om{} and LISA~\cite{lai2023lisa} in Table~\ref{tab:gres}. Three variants of \om{}, including \om{}-Vicuna-7B, \om{}-Vicuna-13B, and \om{}-Llama2-13B show competitive performance without finetuning on gIoU to the strongest non-LLM baseline ReLA~\cite{liu2023gres}. However, LISA models fail to handle the GRES task without finetuning, showing degradation in both gIoU and cIoU in each model variant. Especially the low N-acc indicates that LISA is unable to correctly reject the empty targets. When finetuned on gRefCOCO training set, \om{}-Vicuna-7B performs better than the finetuned LISA counterpart, with about 4\% improvement in gIoU and over 5\% in N-acc on all three evaluation splits. \om{} variants with larger LLM incorporated further push the limits, achieving over 70\% in gIoU on the validation set, 73\% on test set A, and 65\% on test set B. The 13B models also consistently outperform LISA by large margins, demonstrating the superiority of \om{} in GRES task.

\subsection{Referring Expression Segmentation}

\noindent
\textbf{Settings.} To validate the abilities to handle various tasks, we evaluate \om{} in the classic RES task. Following the common evaluation protocols, we test variants of \om{} and LISA equipped with different LLMs on RefCOCO, RefCOCO+~\cite{kazemzadeh2014referitgame}, and RefCOCOg~\cite{mao2016generation}. We follow the UNC split to perform experiments on RefCOCO and RefCOCO+, and UMD split for RefCOCOg. The models are firstly pretrained as in GRES, and then finetuned for 10 epochs with a joint dataset of these three RES training set. The cIoU metric is adopted to measure the model performances.

\noindent
\textbf{Results.} Table~\ref{tab:res} shows the RES results of \om{}. For the pretrained models, all the three variants of \om{} achieve higher cIoU than LISA~\cite{lai2023lisa} by clear margins. Our 7B model outperforms LISA-Vicuna about 2\% cIoU on almost every data split. After finetuning, the preponderance of \om{} over LISA keeps and even enlarges. Specifically, the cIoU metrics of \om{}-Llama2-13B on 8 sets surpass LISA by at least 2.8\%. For the test set of RefCOCOg, the margin even grows to 5.9\%, which exhibits \om{} is also competitive in the classic RES task. 

\begin{table*}
\begin{center}
\setlength{\tabcolsep}{5.5mm}{
\renewcommand\arraystretch{1.0}
\resizebox{\linewidth}{!}{
\begin{tabular}{l|ccc|ccc|cc}
\toprule
\multicolumn{9}{c}{\textbf{Referring Expression Segmentation on RefCOCO, RefCOCO+, and RefCOCOg}} \\
\multirow{2}{*}{Method} & \multicolumn{3}{c|}{RefCOCO (UNC)} & \multicolumn{3}{c|}{RefCOCO+ (UNC)}  & \multicolumn{2}{c}{RefCOCOg (UMD)} \\
& Val. & Test-A & Test-B & Val. & Test-A & Test-B & Val. & Test \\
\midrule
\midrule
MCN~\cite{luo2020multi} & 62.4 & 64.2 & 59.7 & 50.6 & 55.0 & 44.7 &
49.2 & 49.4 \\
VLT~\cite{ding2021vision} & 67.5 & 70.5 & 65.2 & 56.3 & 61.0 & 50.1 & 55.0 & 57.7 \\
CRIS~\cite{wang2022cris} & 70.5 & 73.2 & 66.1 & 62.3 & 68.1 & 53.7 & 59.9 & 60.4 \\
LAVT~\cite{yang2022lavt} & 72.7 & 75.8 & 68.8 & 62.1 & 68.4 & 55.1 & 61.2 & 62.1 \\
ReLA~\cite{liu2023gres} & 73.8 & 76.5 & 70.2 & 66.0 & 71.0 & 57.7 & 65.0 & 66.0 \\
X-Decoder~\cite{zou2023generalized} & - & - & - & - & - & - & 64.6 & - \\
SEEM~\cite{zou2023segment} & - & - & - & - & - & - & 65.7 & - \\
PolyFormer-L~\cite{liu2023polyformer} & 76.0 & 78.3 & 73.3 & \textbf{69.3} & \textbf{74.6} & 61.9 & 69.2 & 70.2 \\
% \midrule
LISA-Vicuna-7B$^*$~\cite{lai2023lisa} & 74.1 & 76.5 & 71.1 & 62.4 & 67.4 & 56.5 & 66.4 & 68.5 \\
\rowcolor{mygray}
\om{}-Vicuna-7B & 76.4 & 77.4 & 72.8 & 64.5 & 67.7 & 58.6 & 71.1 & 72.0 \\
LISA-Vicuna-7B$^*$~\cite{lai2023lisa} (ft) & 74.9 & \textbf{79.1} & 72.3 & 65.1 & 70.8 & 58.1 & 67.9 & 70.6  \\
\rowcolor{mygray}
\om{}-Vicuna-7B (ft) & \textbf{77.2} & 78.9 & \textbf{73.5} & 65.9 & 69.6 & \textbf{59.8} & \textbf{72.7} & \textbf{73.3} \\
\midrule
LISA-Vicuna-13B~\cite{lai2023lisa} & 71.7 & 74.7 & 68.1 & 59.4 & 64.2 & 52.9 & 65.2 & 66.1 \\
\rowcolor{mygray}
\om{}-Vicuna-13B & 74.6 & 77.5 & 70.5 & 62.5 & 66.5 & 55.5 & 69.6 & 71.2  \\
LISA-Vicuna-13B~\cite{lai2023lisa} (ft) & 76.0 & 78.8 & 72.9 & 65.0 & 70.2 & 58.1 & 69.5 & 70.5 \\
\rowcolor{mygray}
\om{}-Vicuna-13B (ft) & \textbf{78.2} & \textbf{80.4} & \textbf{74.2} & \textbf{67.4} & 71.5 & \textbf{60.9} & \textbf{74.2} & \textbf{75.6} \\
\midrule
LISA-Llama2-13B~\cite{lai2023lisa} & 73.4 & 76.2 & 69.5 & 62.3 & 66.6 & 56.3 & 68.2 & 68.5 \\
\rowcolor{mygray}
\om{}-Llama2-13B & 77.7 & 79.9 & 74.2 & 68.0 & 71.5 & 61.5 & 73.2 & 73.9 \\
LISA-Llama2-13B~\cite{lai2023lisa} (ft) & 76.3 & 78.7 & 72.4 & 66.2 & 71.0 & 59.3 & 70.1 & 71.1 \\
\rowcolor{mygray}
\om{}-Llama2-13B (ft) & \textbf{79.2} & \textbf{81.7} & \textbf{77.1} & \textbf{70.3} & 73.8 & \textbf{63.6} & \textbf{75.7} & \textbf{77.0} \\
\bottomrule
\end{tabular}}
}
\end{center}
\vspace{-0.2in}
\caption{Referring expression segmentation results on RefCOCO, RefCOCO+~\cite{kazemzadeh2014referitgame} and RefCOCOg~\cite{mao2016generation} dataset. The cIoU metrics of each split are reported. Baselines are excerpted from~\citet{lai2023lisa}. (ft) denotes the models are finetuned on the joint training set of the referring expression segmentation datasets. * means the results are excerpted from the original paper.
}
\label{tab:res}
\vspace{-0.15in}
\end{table*}

% \vskip -0.2in
\subsection{Referring Expression Comprehension}

\noindent
\textbf{Settings.} Since \om{} is capable for RES tasks, it is natural to transfer to Referring Expression Comprehension (REC) tasks, by simply computing the bounding boxes of the masks. The datasets of REC are the same as RES, in which we evaluate \om{}. If the IoU of a predicted bounding box and the ground truth is greater than 0.5, this prediction is marked as correct. We use the same models in RES to evaluate the phrase grounding capability in REC tasks.

\noindent
\textbf{Results.} As shown in Table~\ref{tab:rec}, we mainly compare our \om{} to LISA in the three variants. Without finetuning, \om{}-Vicuna-7B outperforms LISA with a large margin over 7\% in almost all evaluation sets. The similar trends also hold for the Vicuna-13B and Llama2-13B variants. \om{} also benefits a lot from finetuning, \textit{e.g.,} the 7B variant achieves consistently higher Prec@0.5 than uLLaVA-7B~\cite{xu2023u} with LoRA finetuning, which is another strong baseline that adopts the ``mask2bbox" pipelines without direct bounding box supervision. The finetuned \om{} also shows competitive performances to the fully finetuned uLLaVA-7B, suggesting the strong potential of our method. With larger LLMs incorporated, the performances of \om{} continue with over 3\% increments over all datasets.

\begin{table*}
\begin{center}
\setlength{\tabcolsep}{6mm}{
\renewcommand\arraystretch{1.0}
\resizebox{\linewidth}{!}{
\begin{tabular}{l|ccc|ccc|cc}
\toprule
\multicolumn{9}{c}{\textbf{Referring Expression Comprehension on RefCOCO, RefCOCO+, and RefCOCOg}} \\
\multirow{2}{*}{Method} & \multicolumn{3}{c|}{RefCOCO} & \multicolumn{3}{c|}{RefCOCO+}  & \multicolumn{2}{c}{RefCOCOg} \\
& Val. & Test-A & Test-B & Val. & Test-A & Test-B & Val. & Test \\
\midrule
\midrule
u-LLaVA-7B~\cite{xu2023u} (LoRA) & 83.47 & 87.13 & 80.21 & 68.74 & 76.32 & 60.98 & 76.19 & 78.24 \\
u-LLaVA-7B~\cite{xu2023u} (full-ft) & 86.04 & \textbf{89.47} & 82.26 & 74.09 & \textbf{81.16} & 66.61 & 79.87 & 81.68 \\
% VisionLLM-Intern-H~\cite{wang2023visionllm} & 86.7 & - & - & - & - & - & - & - \\
% \midrule
LISA-Vicuna-7B~\cite{lai2023lisa}                   
& 78.68             & 81.72             & 75.74
& 62.92             & 68.93             & 56.49
& 70.10            & 72.47 \\

\rowcolor{mygray} \om{}-Vicuna-7B    
& 85.50   & 88.01  &  82.49
& 70.21   & 75.62  &  65.11
& 79.00   &  79.21 \\

LISA-Vicuna-7B~\cite{lai2023lisa} (ft) 
& 85.39 & 88.84 & 82.59 
& \textbf{74.23} & 79.46 & \textbf{68.40} 
& 79.34 & 80.42 \\

\rowcolor{mygray} \om{}-Vicuna-7B  (ft)     
& \textbf{86.27} & 89.22 & \textbf{83.77}   
& 72.81 & 78.78 & 68.01 
& \textbf{81.58} & \textbf{81.83} \\
\midrule

LISA-Vicuna-13B~\cite{lai2023lisa}                   
& 80.01 & 83.26 & 76.26 & 63.77 & 70.24 & 57.42 & 71.79 & 73.34 \\
\rowcolor{mygray}
\om{}-Vicuna-13B                  
& 83.12 & 87.01 & 80.54 & 68.14 & 73.90 & 62.00 & 77.08 & 78.89 \\
LISA-Vicuna-13B~\cite{lai2023lisa} (ft)               
& 85.92 & 89.05 & 83.16 & 74.86 &  81.08 & 68.87 & 80.09 & 81.48 \\
\rowcolor{mygray}
\om{}-Vicuna-13B (ft)            
& \textbf{87.71} & \textbf{90.49} & \textbf{84.57} & \textbf{76.52} & \textbf{81.69} & \textbf{70.35} & \textbf{83.90} & \textbf{84.85} \\
\midrule

LISA-Llama2-13B~\cite{lai2023lisa}                   
& 82.52 & 85.56 & 78.82 & 67.91 & 73.77 & 62.25 & 75.37 & 76.83 \\
\rowcolor{mygray}
\om{}-Llama2-13B                  
& 86.99 & 89.54 & 84.08 & 73.89 & 79.10 & 69.38 & 80.68 & 82.07 \\
LISA-Llama2-13B~\cite{lai2023lisa} (ft)               
& 85.91 & 88.84 & 81.73 & 74.46 & 80.56 & 68.26 & 80.09 & 81.27 \\
\rowcolor{mygray}
\om{}-Llama2-13B (ft)                
& \textbf{89.16} & \textbf{92.08} & \textbf{87.17} & \textbf{79.74} & \textbf{84.45} & \textbf{73.41} & \textbf{85.47} & \textbf{86.18} \\
\bottomrule
\end{tabular}}
}
\end{center}
\vspace{-0.2in}
\caption{Referring expression comprehension results on RefCOCO, RefCOCO+~\cite{kazemzadeh2014referitgame} and RefCOCOg~\cite{mao2016generation} dataset. The metric is precision @ 0.5 IoU threshold. (LoRA) means the LLM in u-LLaVA~\cite{xu2023u} is finetuned with LoRA adapter, as LISA and \om{}, while (full-ft) represents the LLM in u-LLaVA is fully trained. Results of u-LLaVA-7B with ``mask2bbox" strategy are reported for fair comparison.}
\label{tab:rec}
\vspace{-0.2in}
\end{table*}

\begin{figure*}
\centering
\includegraphics[width=\linewidth]{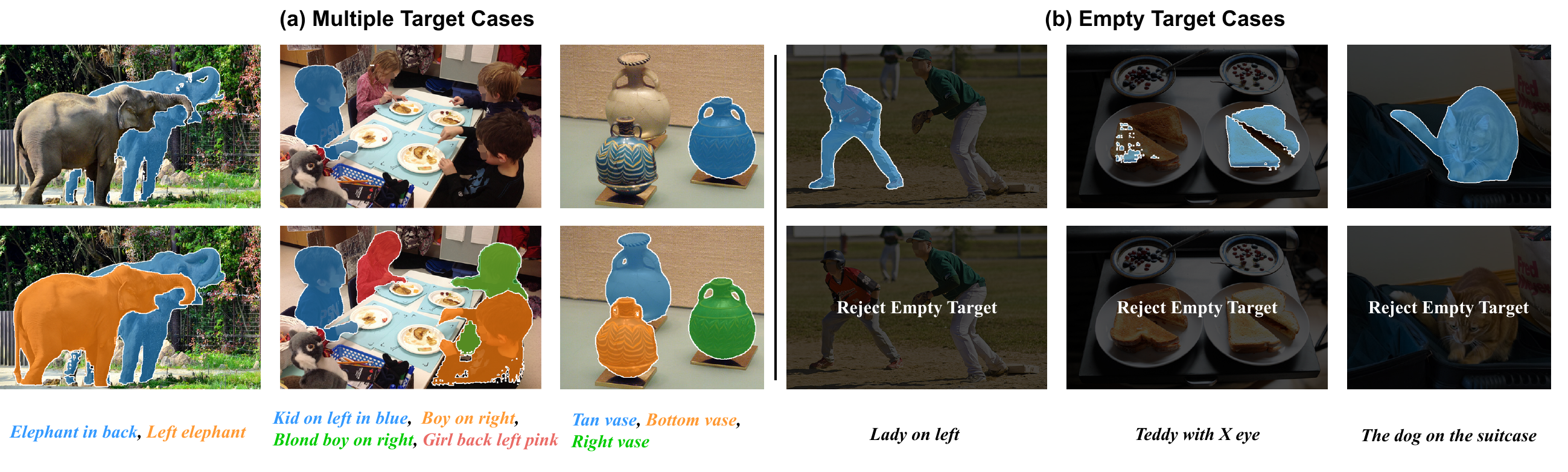}
\vspace{-0.3in}
\caption{Visualizations of \om{} and LISA~\cite{lai2023lisa} in the GRES task. The first row shows LISA's segmentation results, the second row is the masks and rejections of \om{}, and the third row shows the referring expressions in the instructions. In (a) multiple target cases, each target is colored with a specific color. In (b) empty target cases, the images turn darker to highlight the incorrect predictions of LISA. The examples are selected from the gRefCOCO validation set. The masks are generated by the 7B models. Zoom in for the best view.}
\vspace{-0.2in}
\label{fig:fig5}
\end{figure*}

\subsection{Ablation Study}

\begin{table}
\begin{center}
\setlength{\tabcolsep}{2mm}{
\renewcommand\arraystretch{0.9}
\resizebox{\linewidth}{!}{
\begin{tabular}{l|ccc|ccc}
\toprule
\multirow{3}{*}{\tabincell{l}{Model w/ \\ Vicuna-7B}} & \multicolumn{3}{c|}{Modifications} & \multicolumn{3}{c}{gRefCOCO Val.} \\
&
\tabincell{c}{RefExp. \\ +[SEG]}
&
\tabincell{c}{Multiple \\ {[SEG]}}
& 
\tabincell{c}{[REJ] \\ Token}
& gIoU & cIoU & N-acc. \\
\midrule
\om{}                   & \ding{51} & \ding{51} & \ding{51} & 63.32 & 61.70 & 56.45 \\
                        & \ding{51} & \ding{51} & \ding{55} & 51.57 & 60.95 & 30.32 \\
                        & \ding{51} & \ding{55} & \ding{55} & 44.86 & 59.37 & 11.96 \\
LISA~\cite{lai2023lisa}  & \ding{55} & \ding{55} & \ding{55} & 32.21 & 38.72 & 2.71 \\
\midrule
                        & \ding{55} & \ding{51} & \ding{51} & 21.83 & 27.22 & 0.00 \\
\bottomrule
\end{tabular}}
}
\end{center}
\vspace{-0.2in}
\caption{Ablation study on the core designs of \om{}. \ding{51} means the employment of the component while \ding{55} means not. ``RefExp.+[SEG]", ``Multiple [SEG]", and ``[REJ] Token" are short for adding referring expression before [SEG] in the answer prompt, using multiple [SEG] tokens, involving [REJ] token, respectively.}
\label{tab:abl}
\vspace{-0.5cm}
\end{table}

\noindent
\textbf{The involvement of the [REJ] token.} [REJ] token plays a rather important role in \om{}. We study the effect of the [REJ] token by removing it from the vocabulary in \om{}, yielding a variant unable to reject a target from the text outputs. As shown in the 2$^\text{nd}$ row of Table~\ref{tab:abl}, after removing [REJ] token, there is a sharp N-acc drop over 25\% relatively, followed by the decline of gIoU at about 10\% on gRefCOCO validation set. This performance degradation indicates the significance of LLM learning a special token to reject the referred instances absent in the image.

\noindent
\textbf{Learning multiple [SEG] tokens.} We continue to ablate the multiple [SEG] tokens, which is another core design of \om{}. After removing [REJ] token, we then reduce the number of [SEG] tokens to 1, which is identical to LISA~\cite{lai2023lisa} with the referring expression added before the only one [SEG] token: {\small\textit{\textbf{Assistant:} Sure, \{obj\}:[SEG].}}. In the early experiments, we have found that stacking multiple expressions before one [SEG] token would result in divergence. Therefore we separate multiple targets to prompt the model with one expression at a time. The sharp decrements of gIoU by nearly 7\% and N-acc by over 15\% in the 3$^\text{rd}$ row demonstrate the significance of the multiple-[SEG]-token.

\noindent
\textbf{Answers without referring expression.} To examine the efficacy of the hinting prompts, we remove all the referring expressions before the [SEG] tokens. Based on the removal of multiple [SEG]s and [REJ]s, erasing the added referring expression falls back to the original LISA model, as shown in the 4$^\text{th}$ row. We further choose only to remove it from \om{} model, keeping other configurations unchanged. Specifically, if there are two referents, the prompts in the answer will turn to {\small \textit{\textbf{Assistant:} Sure, [SEG], [SEG].}}, whose results are in the last row. The zero N-acc shows the model fails to identify any empty target without the help of the expressions, meanwhile the poor gIoU and cIoU indicates the segmentation ability is damaged. This phenomenon also suggests that the added referring expression hint \om{} to associate each [SEG] token to its corresponding target, which is in coherence with our hypothesis.

\subsection{Visualization}

We visualize some qualitative results of \om{} to verify its effectiveness. As shown in Figure~\ref{fig:fig5}, we present two groups of examples from the validation set of gRefCOCO~\cite{liu2023gres} to see how \om{} outperforms LISA in the face of the two main challenges in GRES: multiple targets and empty targets. In part (a), \om{} has managed to segment all the targets referred to, while LISA could only segment one of the requested instances. For example, in the third column, LISA only predicts the mask of the rightmost vase. On the contrary, \om{} separately segments all three targeted vases. In part (b), LISA mistakenly segments the instances in the image that disagree with the referring expression, e.g., in the sixth column, LISA proposes a mask of a cat in response to the request of the dog on the suitcase, whereas \om{} successfully reject all the empty targets.

%% file: sec/conclusion.tex
\section{Conclusion}
\label{sec:conclu}

% This paper introduces a novel multimodal large language model dubbed Generalized Segmentation Visual Assistant. To tackle the important and challenging generalized referring expression segmentation problem in practical application scenarios, GSVA reuses the [SEG] token and innovatively introduces the [REJ] token, effectively achieving multi-objective segmentation and empty target rejection. Extensive experiments on generalized referring expression segmentation, referring segmentation, semantic segmentation, and other vision language tasks fully demonstrate the superior performance of our method, highlighting its significance for future research and applications.

% pxr: 稍微调整了一下顺序
This paper introduces a novel multimodal large language model dubbed \textbf{\omlong{} (\om{})}. By introducing multiple [SEG] tokens and the new [REJ] token, \om{} effectively achieves multi-objective segmentation and empty target rejection, which addresses the challenging segmentation problems in practical application scenarios, \textit{e.g.}, Generalized Referring Expression Segmentation (GRES). Extensive experiments on GRES, classic RES, and REC tasks fully demonstrate the superior performance of our method, highlighting its significance for future research and applications.

%% file: sec/acknowledgements.tex
\section*{Acknowledgments}
This work is supported in part by the National Key R\&D Program of China under Grant 2021ZD0140407, the National Natural Science Foundation of China under Grants 62321005 and 62276150. We also thank Dr.\@~Yuan~Yao and Prof.\@~Zhiyuan~Liu for their insightful and valuable comments on this research project. We also appreciate their generous support of computational resources.

%% file: sec/discussion.tex
\section{Discussions}

\subsection{Comparison to Concurrent Works}
% \noindent
\textbf{NExT-Chat}~\cite{zhang2023next} decodes boxes and masks from the $\langle\text{trigger}\rangle$ token, showing promising capabilities in many grounded understanding tasks. However, it falls short of explicitly rejecting non-existing objects in user's queries. \textbf{SESAME}~\cite{wu2023see} is adept at correcting the wrong referents and segmenting the closest object in the image by adjusting the input prompt with an alternative to the empty target, while \om{} tackles it with \textbf{[REJ] tokens in a unified output space}. When there are multiple empty targets, \om{} can seamlessly reject them, while SESAME could have undefined behaviors. In summary, the aforementioned concurrent works consider the challenge of multiple and empty targets in the area of segmentation LLMs to a certain extent, while \om{} studies the problem \textbf{more systematically via weight-sharing SEG tokens and the novel REJ token}. 

\subsection{\om{} and Reasoning Segmentation}

\setlength{\columnsep}{0.1in}
\begin{wraptable}{r}{4cm}
\vspace{-0.4cm}
\resizebox{\linewidth}{!}{
\begin{tabular}{l|cc}
Method & gIoU & cIoU \\
\thickhline 
LISA-7B (ft) & 50.5 & 53.2 \\
\om{}-7B (ft) & 50.5 & 56.4 \\
\end{tabular}
}
\vspace{-0.3cm}
\caption{Results of LISA and \om{} on ReasonSeg dataset.}
\label{tab:reasonseg}
\vspace{-0.5cm}
\end{wraptable}

We evaluate \om{} on ReasonSeg~\cite{lai2023lisa} with a generalized configuration. Table~\ref{tab:reasonseg} demonstrates the capability of \om{} to handle instructions with complex logic, showing competitive results to LISA\footnote{Results reproduced by the open-source code of LISA.}. Figure~\ref{fig:greason} exemplifies that \om{} can segment the ReasonSeg referent while rejecting the additional empty target, while LISA fails to make the rejection, which further verifies the efficacy of \om{}.

\subsection{Support of Various Question Types}

\setlength{\columnsep}{0.1in}
\begin{wraptable}{r}{4cm}
\vspace{-0.15in}
\resizebox{\linewidth}{!}{
\begin{tabular}{l|ccc}
Question & gIoU & cIoU & N-acc. \\
\thickhline 
``What" & 63.32 & 61.70 & 56.44 \\
``Where" & 63.43 & 61.57 & 56.98 \\
``Show" & 63.05 & 61.35 & 56.48 \\
``Outline" & 63.16 & 61.51 & 56.27 \\
\end{tabular}
}
\vspace{-0.1in}
\caption{GRES results with different types of questions.}
\label{tab:diff_q}
\vspace{-0.15in}
\end{wraptable}

\begin{figure}
    \centering
    \includegraphics[width=0.9\linewidth]{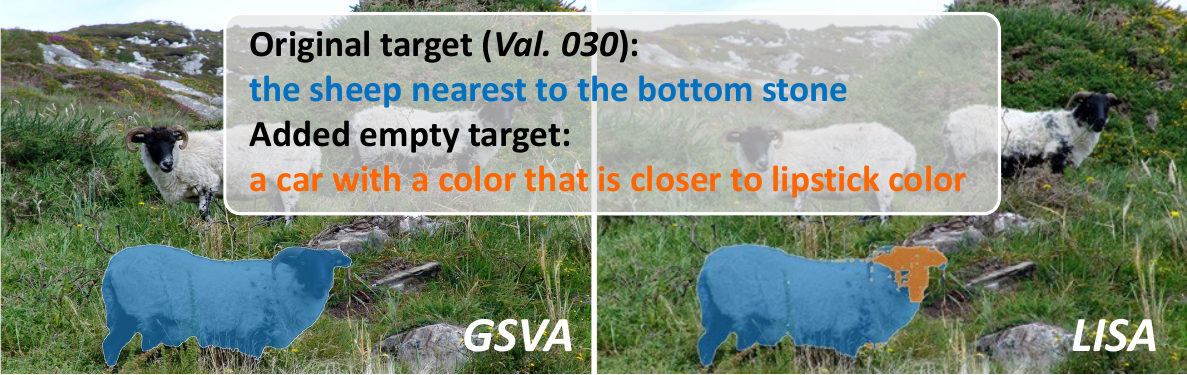}
    \vspace{-0.1in}
    \caption{Generalized Reasoning Segmentation Example.}
    \label{fig:greason}
    \includegraphics[width=0.9\linewidth]{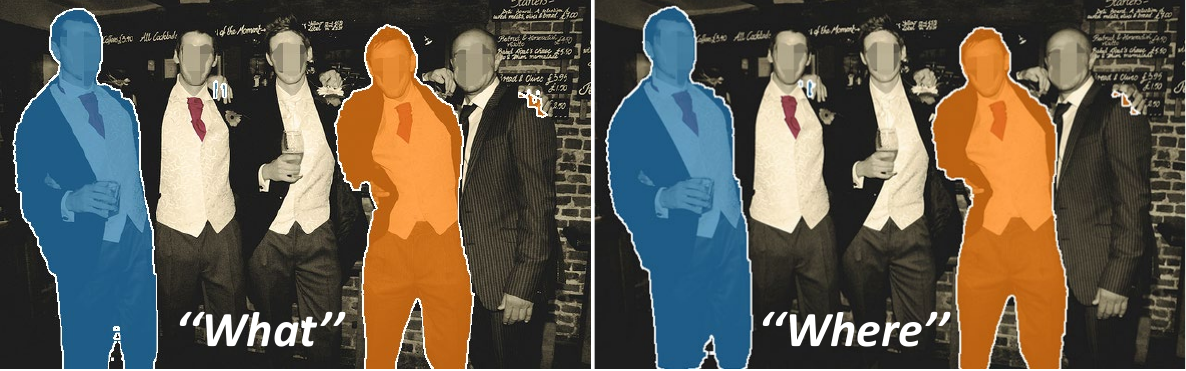}
    \vspace{-0.1in}
    \caption{Example of different types of questions.}
    \label{fig:gprompt}
    \includegraphics[width=0.6\linewidth]{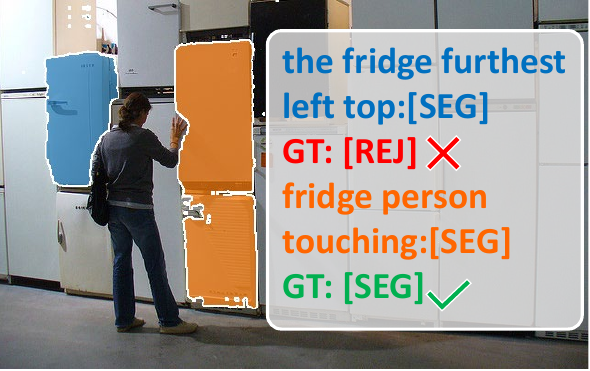}
    \vspace{-0.1in}
    \caption{Failure case example of \om{} in terms of N-acc., where empty targets are incorrectly predicted with [SEG] tokens.}
    \label{fig:fail}
    \vspace{-0.2in}
\end{figure}

For simplicity, we show one type of question in the example in the paper. In contrast, the prompt types are diversified during training, including ``Please segment \{objs\} in this image", "Can you segment \{objs\} in the image", \textit{etc}. Besides, the training data also contains VQA and Reasoning Segmentation, which include various questions and answers. As shown in Table~\ref{tab:diff_q}, apart from the substituted ``Where" question of ``What," we also examine the robustness of \om{} by testing the pretrained model with the unseen ``Show" and ``Outline" questions, short for ``Show me \{objs\} in the image with segmentation masks" and ``Outline \{objs\} in this image with segmentation masks," respectively. As demonstrated in Figure~\ref{fig:gprompt}, different question forms do not impact much performance deviation as long as they are reasonable. 

\subsection{Multiple Objects in One Expression}

In \om{}, each expression is separated with a comma in the prompt, corresponding to a segmentation map or rejection token. If more than one object is stated in the expression, a single [SEG] token will guide a segmentation mask to cover all objects. The common practice of GRES is merging all masks of the objects referred to into one as the ground truth for evaluation, which we follow for fair comparisons. Since there exist numerous expressions containing both present and absent objects, \om{} learns to predict the mask of the union of the referents so the absent objects will not occur.

\subsection{Failure Cases on N-acc.}

Although \om{} achieves a \textbf{state-of-the-art} level of N-acc on the GRES task and outperforms both Non-LLM and LLM models, we still observe that several failure cases are relatively small and vague fragments in the images, easily leading to 
misperception. As shown in Figure~\ref{fig:fail}, there is no fridge obviously at the top left, but the model proposes a nearer fridge instead. It is hard for the model to tell whether in the corner stands a fridge, especially at low input resolution. This suggests using higher-resolution vision encoders, which will be an interesting future direction.

% \subsection{Multiple Empty Targets}

% It is for simplicity that we present a single [REJ] token in Fig.{\color{red}3}(b) in the paper. We would like to emphasize when multiple empty targets are present in question, \om{} predicts a [REJ] token for each empty target since the [REJ] tokens do not decode any segmentation maps. Quantitatively, this design performs well in gRefCOCO dataset where there is usually several empty targets in the referring expression, as shown in Tab.{\color{red}1} of the paper. In addition, the qualitative visualizations in Figure~\ref{fig:vis} also show that \om{} is competent to handle multiple empty targets in complex expressions.

%% file: sec/impl.tex
\section{Implementation Details}

We elaborate on the implementation details of \om{} and the settings of the experiments and list all the hyper-parameters in Table~\ref{tab:impl} for easy reproduction of the results.

\begin{table*}
\begin{center}
\setlength{\tabcolsep}{3mm}{
\renewcommand\arraystretch{1.0}
\resizebox{\linewidth}{!}{
\begin{tabular}{ll|p{0.2\linewidth}<{\centering}p{0.2\linewidth}<{\centering}p{0.2\linewidth}<{\centering}}
\toprule
\multirow{2}{*}{Experiment} & \multirow{2}{*}{Configuration} & \multicolumn{3}{c}{Model} \\
& & \om{}-Vicuna-7B & \om{}-Vicuna-13B & \om{}-Llama2-13B \\
\midrule
\midrule
\multirow{10}{*}{Pretraining} & Dataset Types & \multicolumn{3}{c}{SemSeg, RES, VQA, ReasonSeg} \\
& SemSeg Datasets & \multicolumn{3}{c}{ADE20K~\cite{zhou2019semantic}, COCO-Stuff~\cite{caesar2018cstuff}, Maplilary Vistas~\cite{neuhold2017mapillary}, PACO-LVIS~\cite{ramanathan2023paco}, Pascal-Part~\cite{chen2014detect}} \\
& RES Datasets & \multicolumn{3}{c}{RefCOCO~\cite{kazemzadeh2014referitgame}, RefCOCO+~\cite{kazemzadeh2014referitgame}, RefCOCOg~\cite{mao2016generation}, RefCLEF~\cite{kazemzadeh2014referitgame}, gRefCOCO~\cite{liu2023gres}} \\
& VQA Datasets &  \multicolumn{3}{c}{LLaVA-Instruct-150K~\cite{liu2023llava}} \\
& ReasonSeg Datasets &  \multicolumn{3}{c}{ReasonSeg~\cite{lai2023lisa}} \\
& Epochs / Steps & \multicolumn{3}{c}{50,000 steps, gradient accumulation: 10 steps / update} \\
& Optimizer & \multicolumn{3}{c}{AdamW~\cite{loshchilov2017decoupled}, learning rate: $3\!\times{}\!10^{-4}$, weight decay: 0.0, gradient clip: 1.0} \\ 
& ZeRO~\cite{rajbhandari2020zero} & \multicolumn{3}{c}{Stage: 2} \\
& Batch size & \multicolumn{3}{c}{2 samples / GPU$\times{}$8 GPUs} \\
& LoRA~\cite{hu2022lora} rank & 8 & 64 & 64 \\
\midrule
\multirow{10}{*}{\tabincell{l}{Finetuning on \\ gRefCOCO~\cite{liu2023gres}}} & Dataset Types & \multicolumn{3}{c}{RES} \\
& SemSeg Datasets & \multicolumn{3}{c}{-} \\
& RES Datasets & \multicolumn{3}{c}{gRefCOCO~\cite{liu2023gres}} \\
& VQA Datasets &  \multicolumn{3}{c}{-} \\
& ReasonSeg Datasets &  \multicolumn{3}{c}{-} \\
& Epochs / Steps & \multicolumn{3}{c}{10 epochs, gradient accumulation: 10 steps / update} \\
& Optimizer & \multicolumn{3}{c}{AdamW~\cite{loshchilov2017decoupled}, learning rate: $1\!\times{}\!10^{-4}$, weight decay: 0.0, gradient clip: 1.0} \\ 
& ZeRO~\cite{rajbhandari2020zero} & \multicolumn{3}{c}{Stage: 2} \\
& Batch size & \multicolumn{3}{c}{2 samples / GPU$\times{}$8 GPUs} \\
& LoRA~\cite{hu2022lora} rank & 8 & 64 & 64 \\
\midrule
\multirow{10}{*}{\tabincell{l}{Finetuning on \\ classic RES}} & Dataset Types & \multicolumn{3}{c}{RES} \\
& SemSeg Datasets & \multicolumn{3}{c}{-} \\
& RES Datasets & \multicolumn{3}{c}{RefCOCO~\cite{kazemzadeh2014referitgame}, RefCOCO+~\cite{kazemzadeh2014referitgame}, RefCOCOg~\cite{mao2016generation}, RefCLEF~\cite{kazemzadeh2014referitgame}} \\
& VQA Datasets &  \multicolumn{3}{c}{-} \\
& ReasonSeg Datasets &  \multicolumn{3}{c}{-} \\
& Epochs / Steps & \multicolumn{3}{c}{50,000 steps, gradient accumulation: 10 steps / update} \\
& Optimizer & \multicolumn{3}{c}{AdamW~\cite{loshchilov2017decoupled}, learning rate: $1\!\times{}\!10^{-4}$, weight decay: 0.0, gradient clip: 1.0} \\ 
& ZeRO~\cite{rajbhandari2020zero} & \multicolumn{3}{c}{Stage: 2} \\
& Batch size & \multicolumn{3}{c}{2 samples / GPU$\times{}$8 GPUs} \\
& LoRA~\cite{hu2022lora} rank & 8 & 64 & 64 \\
\bottomrule
\end{tabular}}}
\end{center}
\vspace{-0.2in}
\caption{Detailed configurations and hyper-parameters of \om{} in pretraining and finetuning stages. ``steps" mean the model is trained for given steps, which is implemented by fixing the steps in each epoch and sample data from the datasets, while ``epochs" mean the model is trained across the whole dataset in each epoch.}
\label{tab:impl}
\vspace{-0.1in}
\end{table*}

\noindent
\textbf{Pretraining.} In the pretraining stages, \om{} starts from a pretrained MLLM, \textit{e.g.}, LLaVA-Vicuna-7B~\cite{liu2023llava} and a pretrained SFM, \textit{e.g.}, SAM-ViT-H~\cite{kirillov2023segment}. As shown in Figure 2 in the main paper, the mask decoder $F_\text{mask}$, the segmentation query projector $\psi$, the LoRA~\cite{hu2022lora} adapter weights on query and value projections in the LLM, and the token embeddings in the vocabularies are trainable. During training, the cross entropy language modeling loss on the output text sequence $\tilde{\mathbf{y}}_\text{txt}$ and the ground truth response $\mathbf{y}_\text{txt}$, the output mask $\tilde{\mathbf{y}}_\text{mask}$ is supervised by the combination of a binary cross entropy loss and a DICE loss to minimize the error to the ground truth mask $\mathbf{y}_\text{mask}$. To summarize, the final objective is
\begin{equation}
\begin{aligned}
\mathcal{L}_\text{total}&=\lambda_1\mathcal{L}_\text{LM}(\tilde{\mathbf{y}}_\text{txt}, \mathbf{y}_\text{txt})+\lambda_2\mathcal{L}_\text{BCE}(\tilde{\mathbf{y}}_\text{mask}, \mathbf{y}_\text{mask}) \\
&+\lambda_3\mathcal{L}_\text{DICE}(\tilde{\mathbf{y}}_\text{mask}, \mathbf{y}_\text{mask}),
\end{aligned}
\end{equation}
where $\lambda_1,\lambda_2,\lambda_3$ are the weight of the above losses, following LISA~\cite{lai2023lisa} set to 1.0, 2.0, and 0.5, respectively.

In the pretraining phase, we mainly adopt the mixed dataset configurations in LISA~\cite{lai2023lisa} to mix four types of datasets and sample in training splits by the ratio 9 of Semantic Segmentation (SemSeg), 6 of Referring Expression Segmentation (RES), 3 of Visual Question Answering (VQA), and 1 of Reasoning Segmentation (ReasonSeg). This ratio of 9:6:3:1 is almost kept from LISA for a fair comparison, except for the increment of RES to weight gRefCOCO~\cite{liu2023gres} more. For each task, the sample is uniformly sampled across the dataset. The semantic segmentation datasets consist of ADE20K~\cite{zhou2019semantic}, COCO-Stuff~\cite{caesar2018cstuff}, Mapillary Vistas~\cite{neuhold2017mapillary}, PACO-LVIS~\cite{ramanathan2023paco}, and PASCAL-Part~\cite{chen2014detect}. The RES datasets include RefCOCO, RefCOCO+~\cite{kazemzadeh2014referitgame}, RefCOCOg~\cite{mao2016generation}, RefCLEF~\cite{kazemzadeh2014referitgame}, and gRefCOCO~\cite{liu2023gres}. VQA and reasoning segmentation adopt LLaVA-Instruct-150K~\cite{liu2023llava} and ReasonSeg~\cite{lai2023lisa} as the datasets, respectively.

We train all three model variants (Vicuna-7B, Vicuna-13B, and Llama2-13B) on 8 Tesla A100 GPUs (80GB) for 50,000 steps with a batch size of 2 per device. The learning rate is set to $3\!\times{}\!10^{-4}$ with a gradient accumulation of 10. The weight decay is set to 0, and the gradients are clipped to 1 by the maximum norm. The learning rate is warmed up in 100 steps and linearly decayed. To reduce the GPU memory footprints, we adopt the AdamW~\cite{loshchilov2017decoupled} optimizer with stage 2 ZeRO~\cite{rajbhandari2020zero}. As for LoRA, we pick 8 as the LoRA rank for the 7B variant and 64 for the 13B variants.

\noindent
\textbf{Finetuning on GRES task.} We finetune the pretrained models on gRefCOCO~\cite{liu2023gres} dataset to improve the GRES performances further. For all variants of \om{}, we load the weights of the respective model in pretraining and train another 10 epochs on gRefCOCO. This is slightly different from pretraining because we go through the whole training set once per epoch without sampling data from it. In the finetuning experiments, we use a lower learning rate $1\!\times{}\!10^{-4}$, keeping other configurations unchanged.

\noindent
\textbf{Finetuning on classic RES task.} For classic RES, including RefCOCO, RefCOCO+~\cite{kazemzadeh2014referitgame}, and RefCOCOg~\cite{mao2016generation}, we choose the mixed classic RES dataset during training aside from gRefCOCO. In addition to reducing the learning rate to $1\!\times{}\!10^{-4}$, the other hyper-parameters remain unchanged.

% \noindent
% \textbf{}

%% file: sec/refzom.tex
\section{\om{} on Ref-ZOM}

\noindent
\textbf{Ref-ZOM dataset.} Proposed by DMMI~\cite{hu2023beyond}, Ref-ZOM is a similar dataset to gRefCOCO~\cite{liu2023gres}, posing the challenges of one-to-one, one-to-many, and one-to-zero referring expression to targets in the image relationships in RES. The one-to-one is the case of classic RES where the referring expression is matched with only one target in the image, while the one-to-many and one-to-zero are the multi-target and empty-target cases, respectively. Ref-ZOM contains 55,075 images and 74,942 annotated objects, among which there are 56,972 one-to-one cases, 21,290 one-to-many cases, and 11,937 one-to-zero cases. DMMI provides a default split of training set and test set of Ref-ZOM. There are 43,749 images in the training set and 11,329 images in the test set.

\noindent
\textbf{Setups.} We regard Ref-ZOM~\cite{hu2023beyond} as a kind of GRES, and use a similar protocol to gRefCOCO~\cite{liu2023gres} to evaluate LISA and \om{} on Ref-ZOM dataset. For Ref-ZOM, gIoU and cIoU are substituted to the equivalent metrics, mIoU and oIoU. Different from gRefCOCO, mIoU, and oIoU only count for non-empty targets. For one-to-zero, \textit{i.e.}, empty targets, the accuracy is one if the predicted mask is strictly all-zero. LISA~\cite{lai2023lisa} and our proposed \om{} with Vicuna-7B~\cite{vicuna2023} are evaluated, including the pretrained versions and the finetuned versions with 1 epoch on the training split.

\noindent
\textbf{Results.} As shown in Table~\ref{table:ref_zom}, without finetuning on Ref-ZOM, \om{} surpasses LISA by clear margins in oIoU and mIoU of over 6\%. Also, \om{} approaches close to the SOTA, DMMI~\cite{hu2023beyond}, which is the SOTA proposed along with the Ref-ZOM dataset. After finetuning for 1 epoch, LISA quickly catches up \om{} with less 2\% IoU metrics, while \om{} achieves 94.59\% accuracy in classifying empty targets, keeping a competitive performance to DMMI.

\begin{table}
\begin{center}
\setlength{\tabcolsep}{5mm}{
\renewcommand\arraystretch{1.0}
\resizebox{\linewidth}{!}{
\begin{tabular}{l|ccc}
\toprule
\multirow{2}{*}{Method} & \multicolumn{3}{c}{Ref-ZOM~\cite{hu2023beyond} Test Set} \\
& oIoU & mIoU & Acc \\ 
\midrule\midrule
MCN~\cite{luo2020multi}      & 55.03         & 54.70         & 75.81        \\ 
CMPC~\cite{huang2020referring}       & 56.19         & 55.72         & 77.01        \\ 
VLT~\cite{ding2021vision}        & 60.21         & 60.43         & 79.26        \\ 
LAVT~\cite{yang2022lavt}      & 64.45         & 64.78         & 83.11        \\ 
DMMI~\cite{hu2023beyond}     & \textbf{68.77}& \textbf{68.21}& 87.02 \\
\midrule
LISA-Vicuna-7B~\cite{lai2023lisa} & 60.14 & 61.46 & 72.58 \\
\rowcolor{mygray}
\om{}-Vicuna-7B & 67.12 & 67.98 & 82.66 \\
LISA-Vicuna-7B~\cite{lai2023lisa} (ft) & 66.41 & 65.39 & 93.39 \\
\rowcolor{mygray}
\om{}-Vicuna-7B (ft) & 68.29 & 68.13 & \textbf{94.59} \\
\bottomrule
\end{tabular}}}
\caption{GRES results on the test split of Ref-ZOM~\cite{hu2023beyond} dataset. oIoU and mIoU are only computed on the samples containing targets, while the correct prediction of Acc is the mask of all zeros for empty targets. Baselines are excerpted from ~\citet{hu2023beyond}.}
\label{table:ref_zom}
\vspace{-0.1in}
\end{center}
\end{table}

%% file: sec/semseg.tex
\section{\om{} on Semantic Segmentation}

To verify the vanilla semantic segmentation ability, we include an extra evaluation of the pretrained \om{}-Vicuna-7B and LISA-Vicuna-7B on the ADE20K~\cite{zhou2019semantic} validation dataset. Since the pretraining covers the training set of ADE20K, we only evaluate them to see if they are capable of segmenting semantic regions rather than some specific objects. We prompt the models with the instruction as {\small \textit{\textbf{User:} What is \{classname\} in this image? \textbf{Assistant:} Sure, [SEG].}}, where each \textit{\{classname\}} is filled with the class name that exists in this image. We report the mIoU of the predictions and the ground truths, in Table~\ref{tab:ade}. LISA achieves 60.11 mIoU whereas our \om{} slightly improves to 60.56. 

\begin{table}
\begin{center}
\setlength{\tabcolsep}{5mm}{
\renewcommand\arraystretch{1.0}
\resizebox{0.7\linewidth}{!}{
\begin{tabular}{l|c}
\toprule
Model & mIoU(ss) \\
\midrule
LISA-Vicuna-7B~\cite{lai2023lisa} & 60.11 \\
\om{}-Vicuna-7B & 60.56 \\
\bottomrule
\end{tabular}}
}
\end{center}
\vspace{-0.2in}
\caption{Semantic Segmentation Results of \om{} and LISA~\cite{lai2023lisa} on ADE20K validation dataset. The mIoU is different from the one in the closed-set segmentation, which is computed with the existing classes, and resembles the recall to some extent.}
\label{tab:ade}
% \vspace{-0.2in}
\end{table}

%% file: sec/abl_ana.tex
\section{More Ablation Study}

% \subsection{Ablation Study}

\begin{figure}
    \centering
    \includegraphics[width=0.8\linewidth]{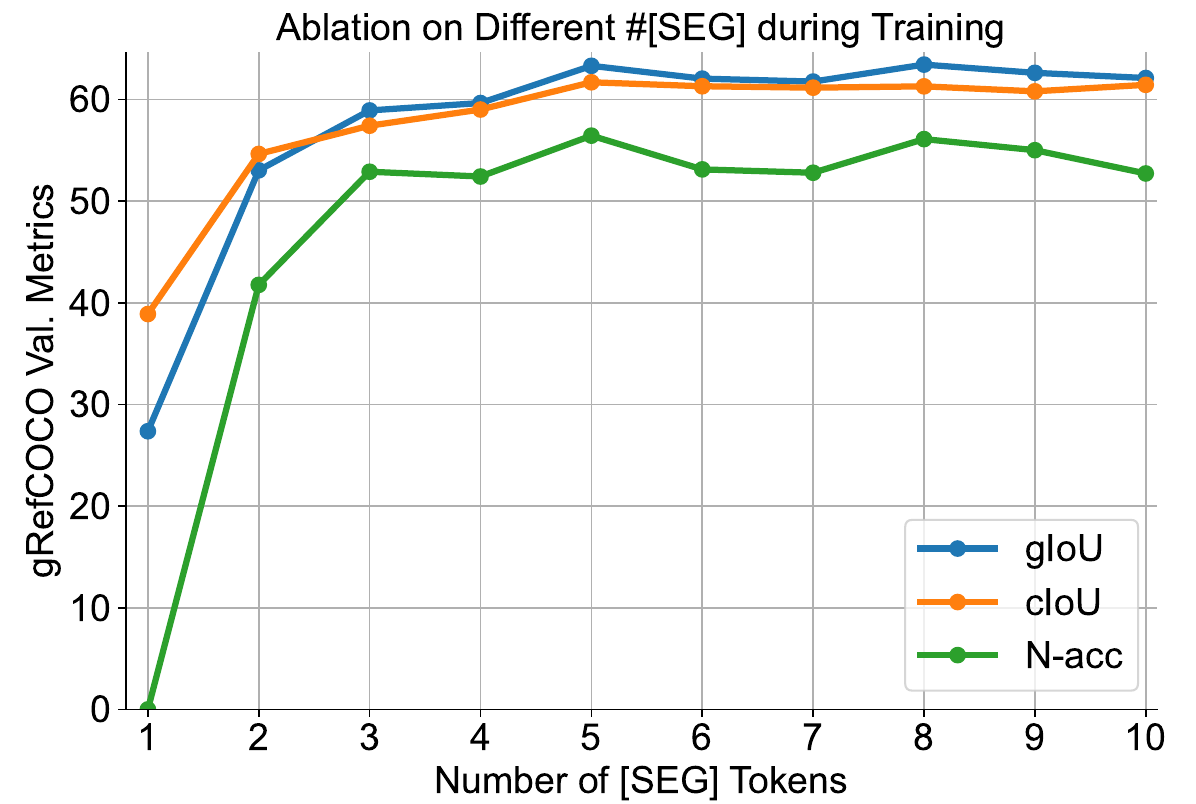}
    \vspace{-0.1in}
    \caption{Ablation on how many [SEG] tokens are used during training. Metrics including gIoU, cIoU, and N-acc on gRefCOCO~\cite{liu2023gres} dataset are reported.}
    \label{fig:abl_num_seg}
    \vspace{-0.1in}
\end{figure}

\noindent
\textbf{The number of targets in training.} We study the effect on how many [SEG] tokens are adopted in training \om{}. In \om{}, the number of [SEG]s controls the number of targets processed by \om{}, which impacts the final capacity of multiple target segmentation. We sweep this hyper-parameter from 1 to 10 and report the results of \om{}-Vicuna-7B on gRefCOCO~\cite{liu2023gres} validation set in Figure~\ref{fig:abl_num_seg}. On the one hand, from the figure, all the metrics include gIoU, cIoU, and N-acc. increases as the number of [SEG] grows from 1 to 5, and the model performances saturate after 5 [SEG] tokens are involved. On the other hand, the memory footprints keep rising as more [SEG] tokens are used and exceed 24GB (RTX3090 / RTX4090) after 6 [SEG] tokens. Considering the performance saturation and the memory consumption, we pick 5 as the default.

\noindent
\textbf{Weight sharing for different [SEG] tokens.} Another design choice considering multiple [SEG] tokens is to use a set of individual [SEG] embeddings rather than sharing weights in \om{}. We adopt 8 tokens from [SEG000] to [SEG007] to study this design. In case of more than 8 targets, we dispatch the targets based on the remainder after divided by 8. However, this design poses convergence difficulties, as shown in Table~\ref{tab:wtshare}. We attribute the intensive performance degradation to the slow convergence of the [SEG] token embedding, which is that too many [SEG] token embeddings could interfere with each other. Therefore, it is reasonable to share weights between [SEG] tokens.

\begin{table}
\begin{center}
\setlength{\tabcolsep}{1.5mm}{
\renewcommand\arraystretch{1.0}
\resizebox{0.85\linewidth}{!}{
\begin{tabular}{l|ccc}
\toprule
\multirow{2}{*}{Model} & \multicolumn{3}{c}{gRefCOCO Val.} \\
& gIoU & cIoU & N-acc. \\
\midrule
\om{} (weight sharing) & 63.32 & 61.70 & 56.45 \\
w/ 8 individual embeddings & 16.13 & 23.11 & 0.00 \\
\bottomrule
\end{tabular}}
}
\end{center}
\vspace{-0.2in}
\caption{Ablation study on weight sharing of [SEG] tokens.}
\label{tab:wtshare}
\vspace{-0.1in}
\end{table}

\noindent
\textbf{Different SAM Backbones.} We alter the default pretrained SAM-ViT-H~\cite{kirillov2023segment} vision encoder to SAM-ViT-L and SAM-ViT-B, to verify the effectiveness of \om{} that is stable to the change of segmentation vision encoder $F_\text{V2}$. As shown in 
Table~\ref{tab:sam}, when we substitute the ViT-H~\cite{dosovitskiy2020image} backbones with SAM pretrained ViT-B/L, the segmentation performances decreases by 4\%$\sim$5\% while the accuracy of empty targets change mildly and even increases to 58.57\% with ViT-B. The slight drops in gIoU and cIoU scores could be attributed to the usages of more miniature vision encoders, and the stable, even higher N-acc implies that generating [REJ] token by MLLM consistently works well with SFMs in different capacities. These results also demonstrate the effectiveness of the design that rejects the empty targets by a special token rather than predicting an all-zero mask.

\begin{table}
\begin{center}
\setlength{\tabcolsep}{2mm}{
\renewcommand\arraystretch{1.0}
\resizebox{0.85\linewidth}{!}{
\begin{tabular}{l|ccc}
\toprule
\multirow{2}{*}{Vision Encoder $F_\text{V2}$ in SFM} & \multicolumn{3}{c}{gRefCOCO Val.} \\
& gIoU & cIoU & N-acc. \\
\midrule
SAM-ViT-H (\om{}) & 63.32 & 61.70 & 56.45 \\
SAM-ViT-L & 61.67 & 60.46 & 56.04 \\
SAM-ViT-B & 59.20 & 56.36 & 58.57 \\
\bottomrule
\end{tabular}}
}
\end{center}
\vspace{-0.2in}
\caption{Ablation study on Different SAM~\cite{kirillov2023segment} ViT backbones.}
\label{tab:sam}
\vspace{-0.2in}
\end{table}

% \subsection{Analysis Experiments}

% \noindent
% \textbf{How each [SEG]/[REJ] token attends to other tokens?}
% \TODO{Attn Vis by Dongchen.}
% \noindent
% \textbf{What does [REJ] token learn?}

%% file: sec/visualizemore.tex
\section{More Visualizations}
We provide more visualization results in Figure~\ref{fig:vis} to further show the effectiveness of our \om{} in GRES task.

\begin{figure*}
    \centering
    \begin{subfigure}[t]{\linewidth}
    \includegraphics[width=0.95\linewidth]{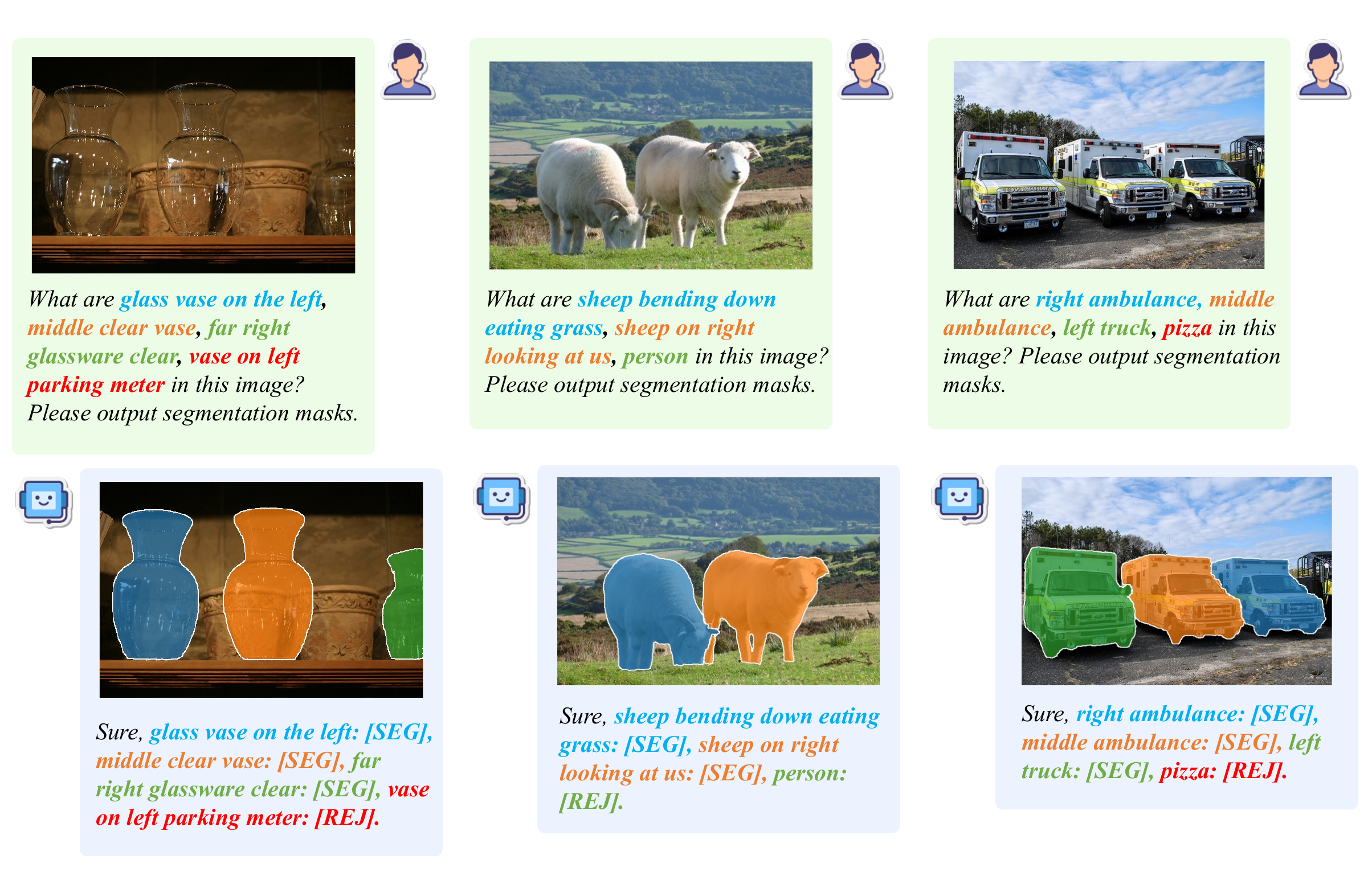}
    \caption{Examples of \om{} on gRefCOCO~\cite{liu2023gres} test set A and B. (\uppercase\expandafter{\romannumeral1})}
    \end{subfigure}
    \begin{subfigure}[t]{\linewidth}
    \includegraphics[width=0.95\linewidth]{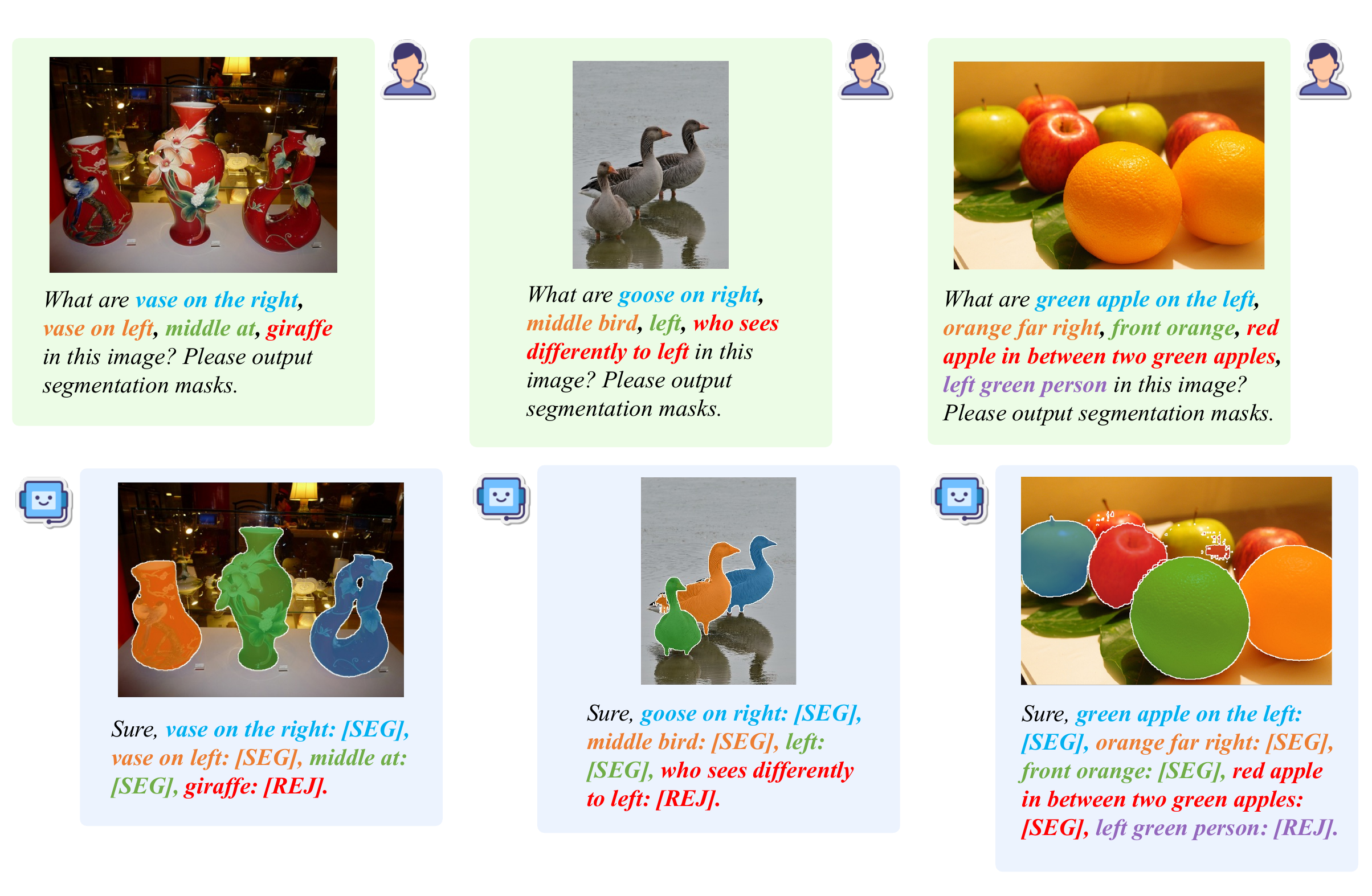}
    \caption{Examples of \om{} on gRefCOCO~\cite{liu2023gres} test set A and B. (\uppercase\expandafter{\romannumeral2})}
    \end{subfigure}
\end{figure*}
\begin{figure*}
    \ContinuedFloat
    \centering
    \begin{subfigure}[t]{\linewidth}
    \includegraphics[width=0.95\linewidth]{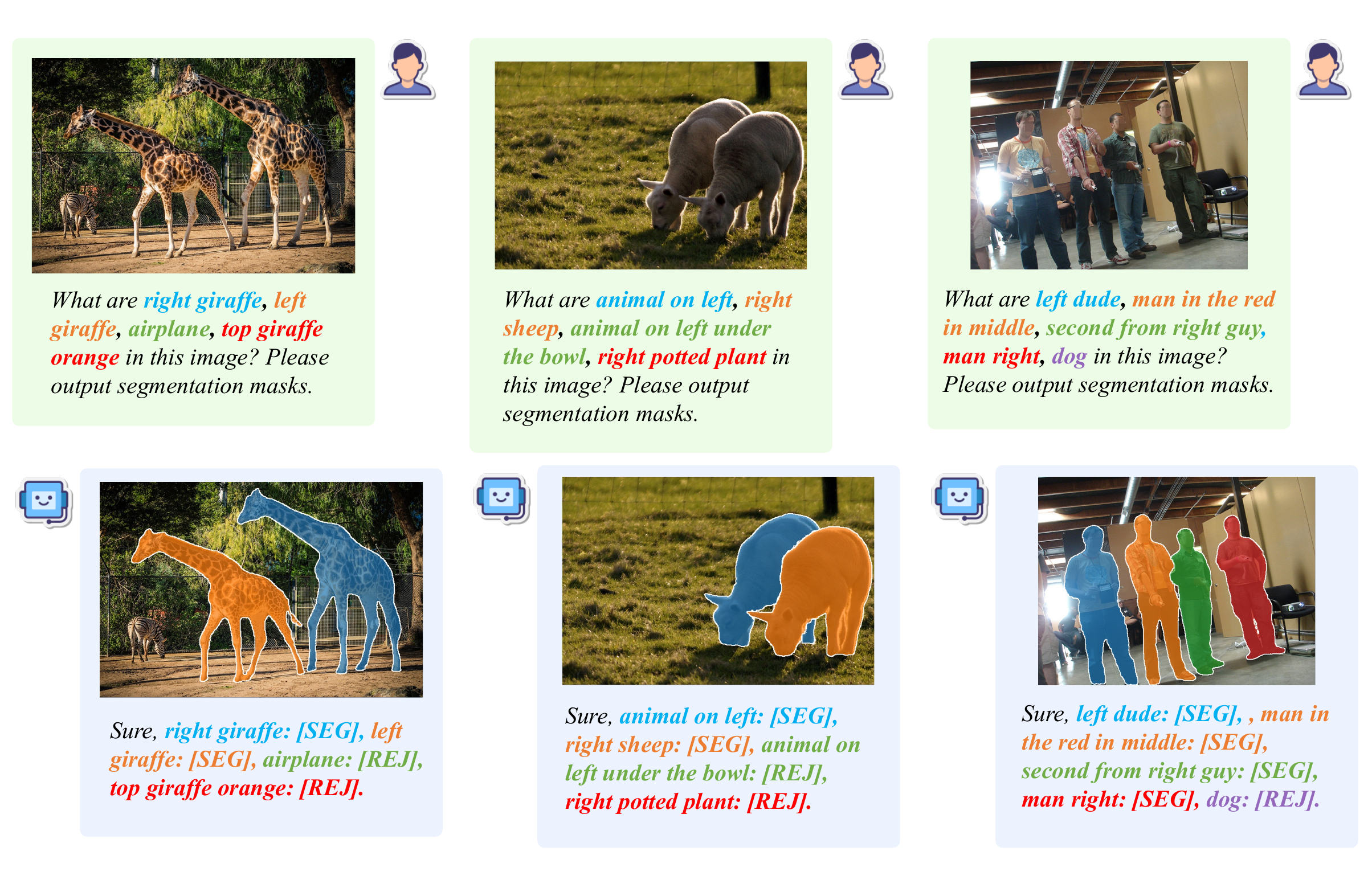}
    \caption{Examples of \om{} on gRefCOCO~\cite{liu2023gres} test set A and B. (\uppercase\expandafter{\romannumeral3})}
    \end{subfigure}
    \begin{subfigure}[t]{\linewidth}
    \includegraphics[width=0.95\linewidth]{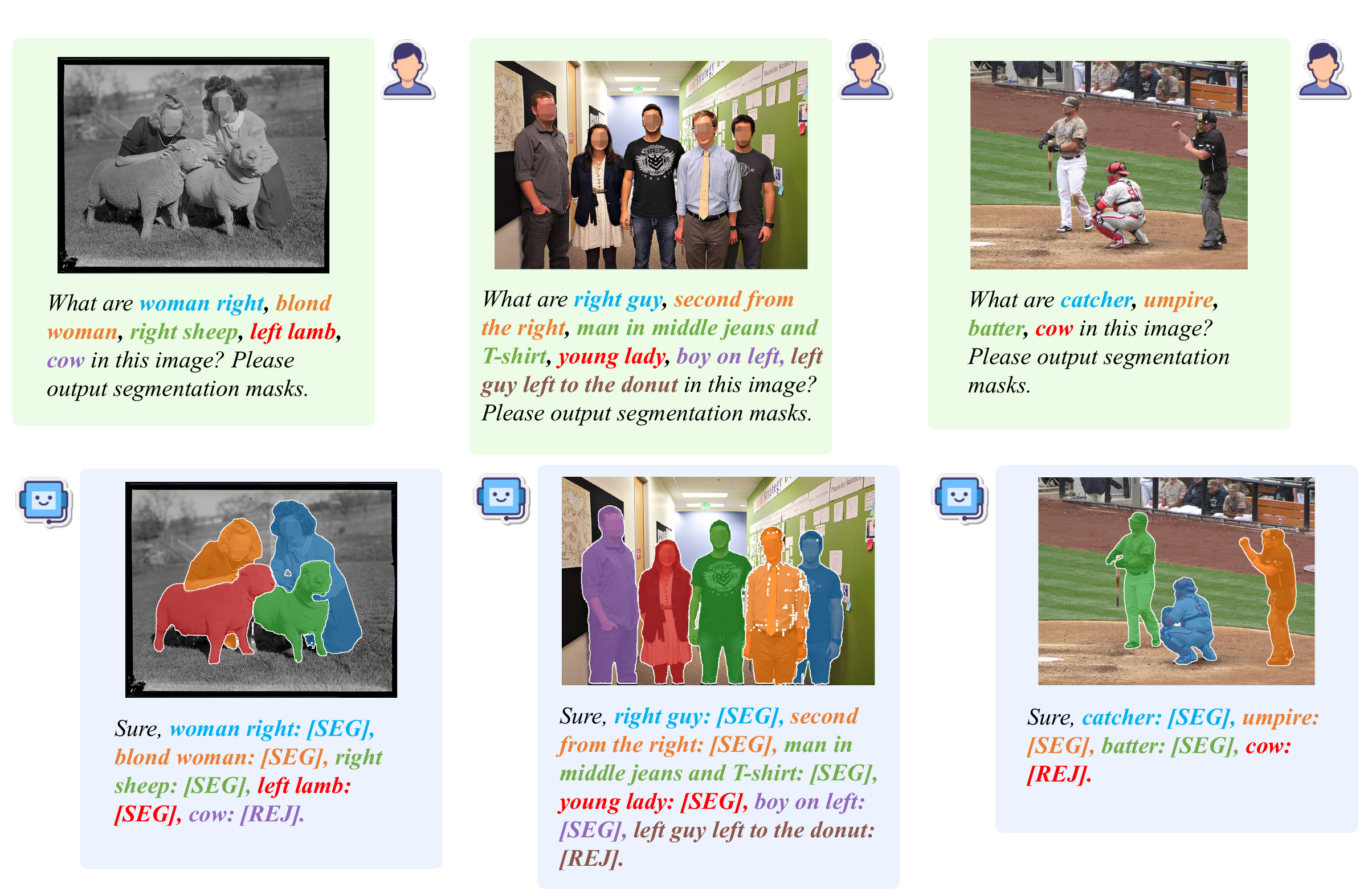}
    \caption{Examples of \om{} on gRefCOCO~\cite{liu2023gres} test set A and B. (\uppercase\expandafter{\romannumeral4})}
    \end{subfigure}
\end{figure*}
\begin{figure*}
    \ContinuedFloat
    \centering
    \begin{subfigure}[t]{\linewidth}
    \includegraphics[width=0.95\linewidth]{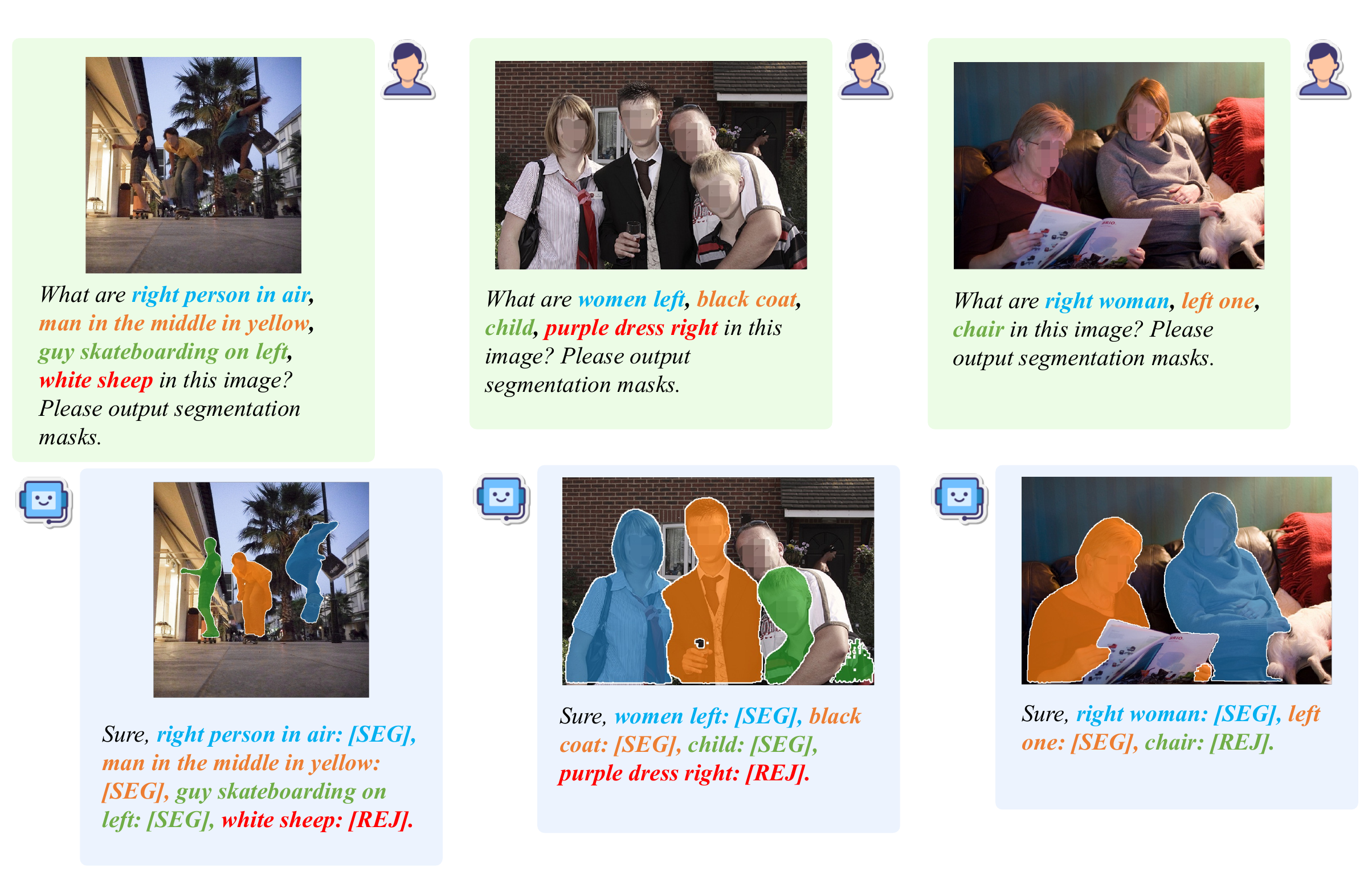}
    \vspace{-0.1in}
    \caption{Examples of \om{} on gRefCOCO~\cite{liu2023gres} test set A and B. (\uppercase\expandafter{\romannumeral5})}
    \end{subfigure}
    \begin{subfigure}[t]{\linewidth}
    \centering
    \includegraphics[width=0.65\linewidth]{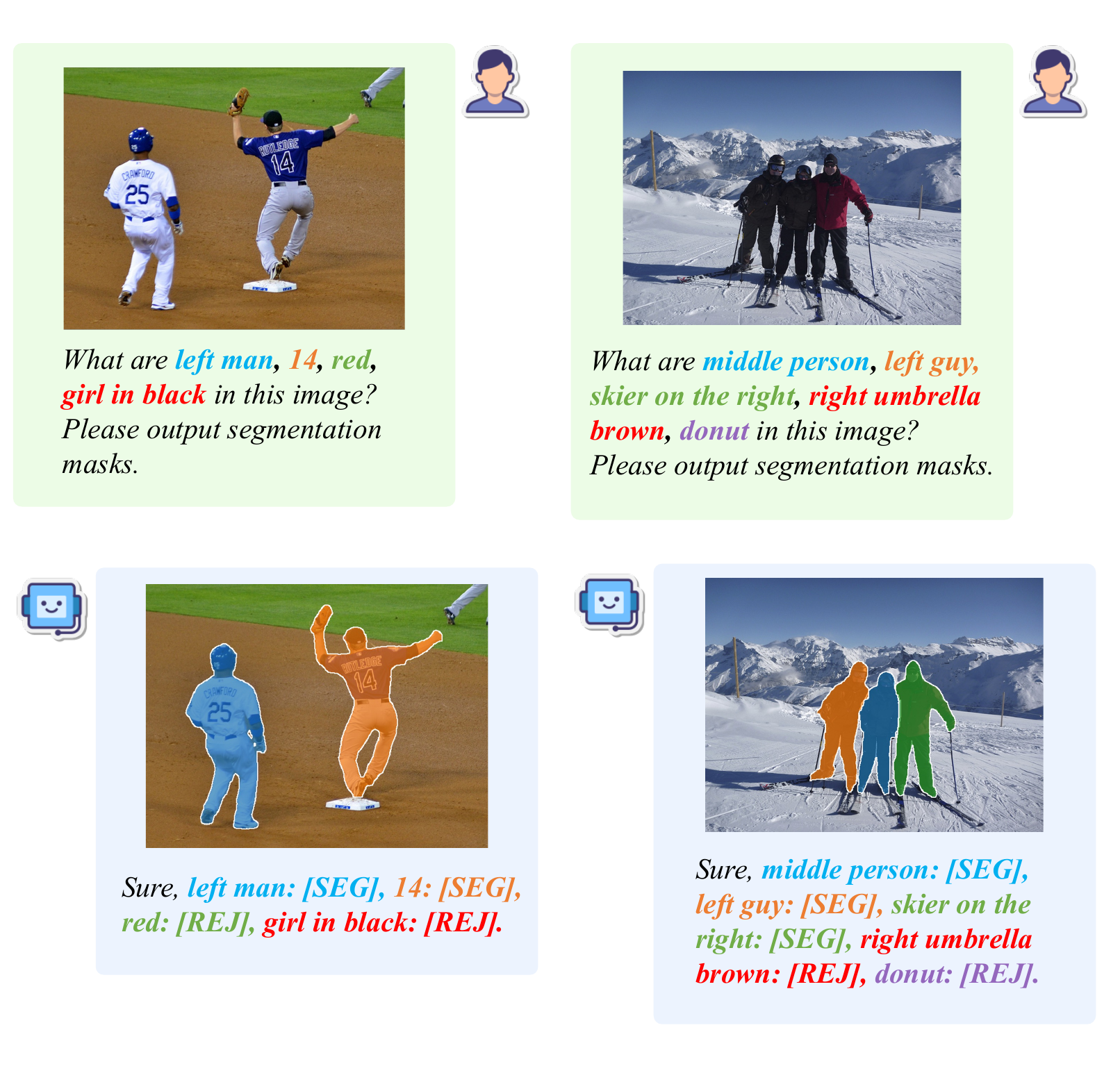}
    \vspace{-0.2in}
    \caption{Examples of \om{} on gRefCOCO~\cite{liu2023gres} test set A and B. (\uppercase\expandafter{\romannumeral6})}
    \end{subfigure}
    \vspace{-0.21in}
    \caption{Visualization of \om{} in GRES task, the inputs and outputs are presented in the form of dialogues between human user and the chatbot. The examples are selected from gRefCOCO~\cite{liu2023gres} test set A and B. Zoom in for best view.}
    \label{fig:vis}
\end{figure*}